\documentclass[]{informs-ijds} 

\OneAndAHalfSpacedXII 

\usepackage{natbib}
\bibpunct[, ]{(}{)}{,}{a}{}{,}%
\def\bibfont{\small}%
\TheoremsNumberedThrough 

\EquationsNumberedThrough 

\usepackage{times,amsmath,epsfig, amssymb} 
\usepackage{graphicx,tabularx}
\usepackage{url}
\usepackage{multirow}
\usepackage{setspace}
\newcommand{\SAVE}[1]{}

\usepackage{paralist}

\usepackage{algorithm}
\usepackage{algorithmic}
\usepackage{longtable}
\usepackage{color}
\usepackage{hhline}

\usepackage{amsmath}
\usepackage{etoolbox}
\usepackage{bm}
\usepackage{mathtools}
\usepackage{algorithm}
\usepackage{algorithmic}
\usepackage{subfigure}
\usepackage{xspace}

\newcommand{\MyMapTemplatePrefixc}[4]{\expandafter#1\csname#3#4\endcsname{#2{#4}}} 
\forcsvlist{\MyMapTemplatePrefixc {\def} {\mathcal}{c}} {A,B,C,D,E,F,G,H,I,J,K,L,M,N,O,P,Q,R,S,T,U,V,W,X,Y,Z}  

\newcommand{\MyMapTemplatePrefixtb}[5]{\expandafter#1\csname#4#5\endcsname{#2{#3{#5}}}} 
\forcsvlist{\MyMapTemplatePrefixtb {\def} {\tilde}{\mathbf}{t}} {A,B,C,D,E,F,G,H,I,J,K,L,M,N,O,P,Q,R,S,T,U,V,W,X,Y,Z}  
\forcsvlist{\MyMapTemplatePrefixtb {\def} {\tilde}{\mathbf}{t}} {0,1,a,b,c,d,e,f,g,h,i,j,k,l,m,n,o,p,q,r,s,t,u,v,w,x,y,z}  

\newcommand{\MyMapTemplateNoPrefix}[3]{\expandafter#1\csname#3\endcsname{#2{#3}}}
\forcsvlist{\MyMapTemplateNoPrefix {\def} {\mathbf} } {0,1,a,b,c,d,e, f, g, h, i, j, k, l, m, n, o, p, q, r, s, t, u, v, w, x, y, z} 
\forcsvlist{\MyMapTemplateNoPrefix {\def} {\mathbf} } {A,B,C,D,E,F,G,H,I,J,K,L,M,N,O,P,Q,R,S,T,U,V,W,X,Y,Z}  

\def\ga{\alpha}
\def\gb{\beta}

\def\gt{\theta}

\begin{document}



\RUNTITLE{Modeling Financial Products}

\TITLE{Modeling Financial Products and their Supply Chains
\footnote{Data Ethics Note:No data ethics considerations are foreseen related to this paper.}}

\ARTICLEAUTHORS{%

\AUTHOR{Margr\'et V. Bjarnad\'ottir}
\AFF{University of Maryland \EMAIL{mbjarnad@umd.edu}, \URL{}}

\AUTHOR{Louiqa Raschid}
\AFF{University of Maryland, \EMAIL{lraschid@umd.edu}, \URL{}}

} 

\ABSTRACT{%

The objective of this paper is to explore how financial big data and machine learning methods 
can be applied to model and understand financial products. 
We focus on residential mortgage backed securities, \texttt{resMBS}, which were at the heart 
of the 2008 US financial crisis. 
These securities are contained within a prospectus and have a complex waterfall payoff structure.
Multiple financial institutions form a supply chain to create prospectuses. 
To model this supply chain, we use unsupervised probabilistic methods, particularly 
dynamic topics models (DTM), to extract a set of features (topics) reflecting community 
formation and temporal evolution along the chain. 
We then provide insight into the performance of the \texttt{resMBS} securities and the impact of 
the supply chain through a series of increasingly comprehensive models. 
First, models at the security level directly identify salient features of \texttt{resMBS} securities 
that impact their performance. 
We then extend the model to include prospectus level features and 
demonstrate that the composition of the prospectus is significant.
Our model also shows that communities along the supply chain that are associated with the 
generation of the prospectuses and securities have an impact on performance.
We are the first to show that toxic communities that are closely linked to financial institutions that
played a key role in the subprime crisis can increase the risk of failure of \texttt{resMBS} securities.
}%


\HISTORY{}

\maketitle


%



\section{Introduction.}\label{intro} 

The 2008 US financial crisis highlighted several challenges and limitations 
in monitoring systemic risk. Informally, systemic risk corresponds to risks 
that impact multiple systemically important financial institutions, 
as was the case in 2008. The concept also includes situations where the risk may be 
contagious across multiple financial markets or institutions. 
The residential mortgage backed securities, \texttt{resMBS}, that were at the heart of the 2008 crisis
were linked to massive defaults of subprime mortgages in the residential mortgage market. 
The risk was contagious due to the interconnectedness of \texttt{resMBS} 
securities with other financial products. For example, a \texttt{resMBS} security 
may have been one arm of a collateralized debt obligation (CDO), while the other 
arm of the CDO could have leveraged a security tied to a foreign currency exchange rate. 
The crisis also exposed a major shortcoming: the gaps in data collection 
around financial products and the limited ability for synthesis across datasets. 
With multiple financial regulatory agencies each responsible for monitoring individual lines of 
financial products, silos of isolated data developed, resulting in a lack of synthesis and analysis 
across products and markets. 
These silos and gaps severely restricted the ability of regulators to manage 
the crisis with data driven actions.

The primary villains of the 2008 crisis were generally believed to be the financial institutions 
that were closely linked to the subprime residential mortgage market, and
that issued subprime loans to borrowers with poor credit scores. 
Many of these institutions filed for bankruptcy protection either before or during the 2008 crisis, 
or they faced financial penalties. 
The rating agencies were also vilified for assigning the coveted \texttt{Aaa} rating to 
many securities, including those 0 with subprime mortgages. 
However, an ex-post study conducted 10 years after the crisis by Ospina and Uhleg 
(\cite{Ospina2018}) painted a more nuanced picture. 
Their research concluded that the {\em majority} of the \texttt{resMBS} securities 
actually performed well. As expected, securities issued closer to the 
financial crisis, i.e., those issued from 2006 to 2007, suffered greater losses than 
those issued before the crisis, but it was only a {\em small share} of \texttt{resMBS} 
securities that incurred the greatest losses. 

Given this scenario and the many challenges of monitoring these products, we believe that 
it would be invaluable for regulators and analysts to have access to the data and models 
that they need in order to gain insight into the systemic risk associated with these products. 
A valuable strategic outcome of our research would be the {\em ability to identify the 
subset of securities} that were found to pose the greatest risk (\cite{Ospina2018}). 
In this study, we use information extraction and integration approaches 
and machine learning methods to address this challenge. 
This study also provides a proof of concept of how novel datasets 
and models that are built on a rich set of features 
can be synergistically exploited to help us understand financial products. 

A key contribution of our research is an understanding of the impact of the supply chain; 
this impact has not been previously studied. 
During the product development of \texttt{resMBS} securities, many financial institutions 
come together to create communities. 
These various institutions perform tasks including originating residential mortgages, 
creating a real-estate trust or legal framework, issuing prospectuses that define the 
contractual obligations of these products, selling securities to institutional investors, 
and finally servicing the \texttt{resMBS} securities, i.e., managing the payouts. 
Forming these communities can incur a significant overhead.
Consequently, a successful community may benefit from staying together and may 
continue to repeatedly issue hundreds of prospectuses over time.
Our study is the first to extract and exploit this community structure. 
{\em We hypothesize that identifying communities could enable us to potentially isolate the 
subset of prospectuses and securities that contributed to the greatest losses.}

We study the performance of \texttt{resMBS} securities using two synergistic steps.
First we model the supply chain using unsupervised learning. 
We draw upon probabilistic topic models (\cite{BNJ2003,SG2007}) that have been 
successfully used to identify topics from a collection of documents. 
In our context, the topics represent communities of financial institutions that work 
together along the supply chain. 
We identify communities that capture complex relationships, e.g., 
\texttt{ Company X is the issuer of securities where Company Y originated the mortgages, 
and X issued prospectuses from 2002 through 2005.} We use dynamic topic models (DTM) 
(\cite{BleiDTM2006}) that use time slices to understand the evolution of topics 
(i.e., communities) over time. 

Second, we apply a series of increasingly comprehensive models to study the financial 
performance of \texttt{resMBS} securities. 
First, models at the level of individual securities identify the salient features that 0
t their financial performance. 
This includes the security class (from secured debt to unsecured debt), the year of issuance, 
and the Moody's initial rating (MIR). We further account for the significant security level 
features reflecting the often complex waterfall payoff structure for these prospectuses
(as detailed in Section \ref{sec:related}). 
For example, the payoff to a security may be subordinated to the payoff to other senior securities;
this is captured by a feature labeled \texttt{senior subordinated debt (SSUP)}.

Our models reveal that the initial ratings may only be able to capture some of the 
nuanced risk arising from the complex waterfall payoff structure. 
Given this complex structure of payouts, it can be hypothesized that the
performance of a security may be influenced by the structure of the prospectus. 
We therefore extend the model to include prospectus level features. 
This includes composition features, e.g.,
the count and the nominal principal (in USD) of the securities in the different security classes, 
within a prospectus.
Our model is the first to demonstrate that the composition of security classes within a prospectus 
can provide a quality signal. Further, the {\em act of including} securities labeled
\texttt{senior subordinated debt (SSUP)} within the prospectus has an impact on performance,
both for the securities that carry the label and, more notably, for other securities within the 
prospectus that do not carry this label.

Our final and most comprehensive model includes the community (topic) associated with each prospectus.
This model highlights the fact that {\em even after accounting for detailed security and 
prospectus level features, 
the community remains significant} in predicting financial performance. 
This is especially notable since we do not include any features of the securities or prospectuses,
or financial performance features,  
during the process of identifying financial communities, 

We use real-world evidence to identify (toxic) financial institutions that were 
active in the subprime market and failed or faced significant difficulty. 
Based on the financial institutions that are prominently linked to the communities, 
we label the communities as toxic, partially toxic or non-toxic.
We compare toxic topics versus all topics with respect to the significance of topics 
in our comprehensive model. 
We demonstrate that toxic topics, and some partially toxic topics, increase the risk of failure, 
while all of the non-toxic topics decrease the risk of failure. 
This notable and novel finding provides strong support for our insight that the 
financial supply chain should be used as a feature in predicting performance.
A further examination of partially toxic topics provides nuanced insights, e.g., 
we can identify when some features of the prospectuses or securities within these topics 
can increase the risk of failure.

We note that we are not currently claiming a causal connection between the 
toxic supply chain communities (toxic topics) and the performance of \texttt{resMBS} securities. 
We well understand that the supply chain is only one of myriad factors that can influence 
the performance of these complex financial products (as our models highlight). 
We further acknowledge that our retrospective data is not free from endogeneity concerns 
(including temporal dependencies), and we explain in detail how we address this issue.
Nevertheless, our novel approach to modeling these products may prove to be groundbreaking, as it shows the synergistic potential of utilizing financial big data and computational methods 
to understand the performance of financial products.


\section{Background and Related Work}\label{sec:related}
\subsection{Background}
\begin{figure}[h]
\begin{center}
\includegraphics[width=8cm]{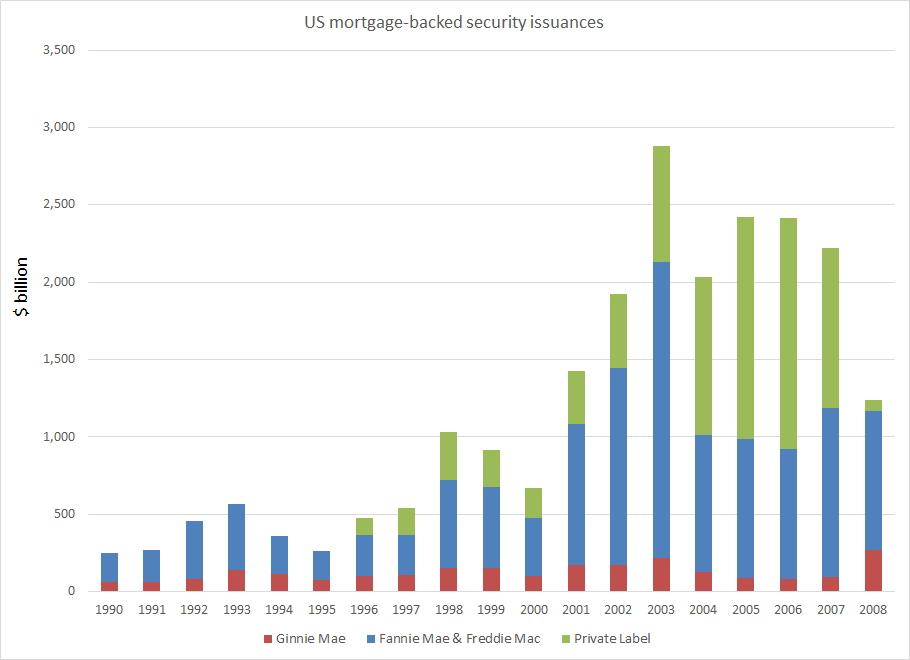}
\end{center}
\begin{center}
\caption{The nominal value of residential mortgage backed securities issued
from 1990 to 2008 
(Securities Industry and Financial Markets Association (\cite{SIFMA} (2018)).}
\label{fig:all_mbs}
\end{center}
\end{figure}

Mortgage backed securities have had a long and successful history as versatile financial 
products and investment vehicles. 
They have been issued since 1990 by various government sponsored enterprises (GSEs), such 
as Fannie Mae, and by non-agency or private labels. 
Our focus is on the private label securities, since they played a major role in the 2008 crisis. 
Figure \ref{fig:all_mbs} provides the nominal value associated with such issuance from 
1990 through 2008 (Securities Industry and Financial Markets Association (\cite{SIFMA} (2018)). 
As reflected in the figure, the early 2000s witnessed a boom in the issuance of 
private label \texttt{resMBS} products. 
The issuance reached its peak around 2006, followed by a sharp decline in late 2007, 
in alignment with the financial crisis. 

\begin{figure}[h]
\begin{center}
\includegraphics[width=10cm]{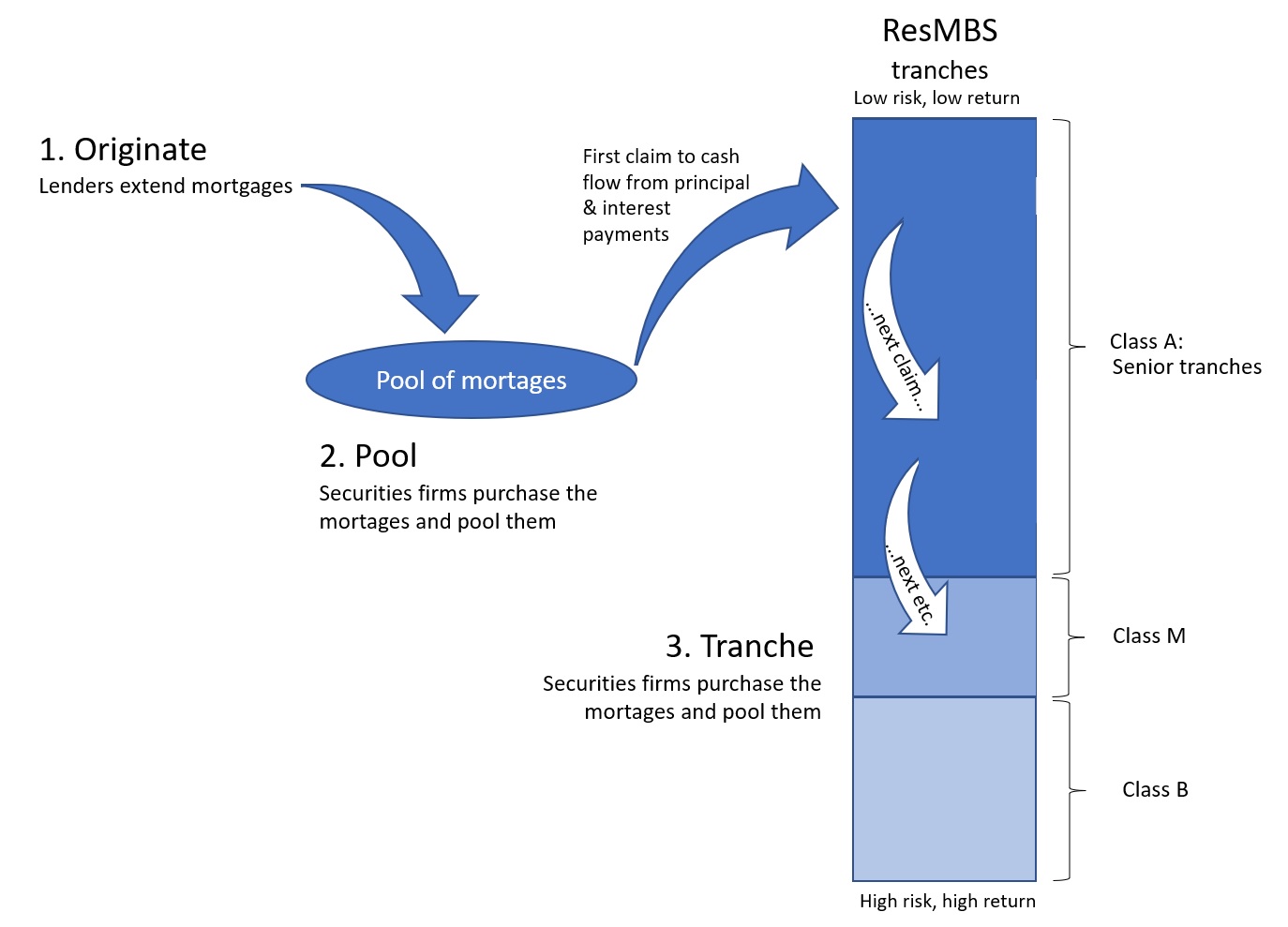}
\end{center}
\begin{center}
\caption{The complex tranche and waterfall payoff structure of residential mortgage backed securities \texttt{resMBS}, adapted and simplified from (\cite{TFCIC2011}).}
\label{fig:rmbs_waterfall}
\end{center}
\end{figure}

A residential mortgage backed security is a financial product constructed 
by pooling cash flows from a collection of residential mortgage loans. 
The cash flows from the pool of mortgages, both the principal and interest payments, 
are allocated to a set of \texttt{resMBS} securities, 
based on a complex waterfall payoff priority. 
Figure \ref{fig:rmbs_waterfall} provides an illustration of the pool of mortgages 
and the waterfall payment structure (adapted from \cite{TFCIC2011}).
The securities served first, at the top of the waterfall, are referred to as 
investment grade senior secured debt, or Class \_A\_ securities. 
This is followed by mezzanine (Class \_M\_) securities, and then discounted or 
unsecured securities (Class \_B\_) securities. 
The securities are packaged within a prospectus – a legal contract – and rated for their credit worthiness. 
The prospectus is issued and the securities are then sold to investors.

\subsection{The 2008 US Financial Crisis and Financial Monitoring}

Since the 2008 financial crisis, several papers have described \texttt{resMBS} 
securities and their role in the crisis. 
We refer the reader to studies by \cite{AS2008,AGV2010,GLSW2010,Hunt2014,Levitin2012}, 
which explicitly 
address how the residential mortgage registration and transfer systems and the subprime crisis 
contributed to the 2008 US financial crisis. 

Researchers have examined the performance of the \texttt{resMBS} securities from different angles, 
including the variation in their credit rating over time (\cite{ AGV2010}),
the effect of issuer size on initial prices (\cite{He2012}) and ratings (\cite{He2011}),
and the role of accounting rules on \texttt{resMBS} sales (\cite{Merrill2012}). 
Levitin and Wachter argue that information failure caused by the complexity and heterogeneity of 
private-label \texttt{resMBS} securities contributed to the crisis, and they propose
that mortgage securitization and origination be standardized as a way of reducing 
complexity and heterogeneity (\cite{Levitin2012}).

We briefly report on the waterfall structure and the ratings of securities that are
issued between 2001 and 2007 (\cite{AGV2010}). 
These authors report that the fraction of highly rated securities in each prospectus decreases with 
increasing mortgage credit risk (measured either ex-ante or ex-post).
This suggests that ratings indeed contain useful information for investors. 
However, there is also evidence of significant time variation in credit ratings, 
including a progressive decline in maintaining strict rating standards between 2005 and 2007.
We note that this was the peak period for the issuance of low performing 
(toxic subprime) mortgages.
The researchers observed high mortgage defaults and losses and large rating downgrades 
for securities with observably higher risk mortgages based on a simple ex-ante model. 
They also observed securities with a high fraction of opaque (low documentation) loans. 
These findings held over the entire sample period. 

An ex-post study by Ospina and Uhleg (\cite{Ospina2018}) provides a detailed picture of the 
payoff performance of the \texttt{resMBS} securities. 
The authors found that the mis-rating of \texttt{Aaa} debt was modest; 
most notably, only a {\em small share} of \texttt{resMBS} securities 
contributed the greatest losses. 
They further concluded that after controlling for the decline in home prices, 
the boom in the housing market was beneficial for the performance of 
the \texttt{resMBS} securities.

In contrast to this previous research, our research focuses on financial institutions 
and their role(s) in the supply chain. 
A single study (\cite{Demiroglu2012}) focuses on the effect when a particular financial 
institution is both an originator of mortgages and either a sponsor or 
a servicer; in these cases, the default rates are significantly lower. 
No other studies attempt to identify and map the supply chain behind 
\texttt{resMBS} products or to estimate its impact. This study fills that gap
and our results suggest a more nuanced scenario where financial communities issue 
prospectuses and securities that have differentiated performance.

Since the 2008 US financial crisis, there has been research on monitoring for systemic risk, 
with a focus on systemic shocks, financial stability, and capital reserves, among other concerns 
(\cite{Arnold2012SystemicRM,Bisias2012,Liang2013}).
In a seminal study, Billio et al. create a novel dataset to study systemic risk 
(\cite{Billio2012}). 
They explore the interconnectedness of hedge funds, banks, brokers, and insurance companies 
using Granger causality. Our research is similar in spirit, but we apply topic modeling to 
study the interconnectedness among financial entities in the \texttt{resMBS} product supply chain. 
We believe that we are the first to apply topic modeling in this context.

\subsection{Text Analytics in Finance}

Our focus in this paper is to understand the drivers of the performance of financial 
products, including the impact of the {\em financial communities} that support the 
\texttt{resMBS} supply chain. This includes a financial institution’s participation in a community and the role(s) it plays. 
To achieve this goal, we draw upon extensive research on document corpora 
in the reference disciplines of information retrieval, computational linguistics and 
machine learning. We further utilize text analytics and topic modeling. 

In this pursuit, we build on a long history of applying text analytics to understand firm behavior 
and financial product performance. 
Loughran and McDonald (\citeyear{LM2016}) offer an extensive survey of 
text analytics approaches in the accounting, finance, and economics literature. 
Their survey covers methods to determine topics in documents, 
to find hidden structures and to determine sentiment. 
As an example of text mining applications in finance, Hoberg and Philips (\citeyear{HP2018})
study a collection of 10-K statements \footnote{10-K statements are comprehensive financial reports that 
publicly traded companies are required to produce.} 
filed with the Securities and Exchange Commission (SEC) to determine how firms 
utilize product descriptions and product differentiation in comparison to their competitors. 
As a result of the study, the authors provide a new set of industries, competitor sets, and 
corresponding new measures of industry competition. 

There is growing interest in exploiting financial big data and computational tools to better model 
and predict the behavior of financial ecosystems. The success of text extraction tools 
(\cite{burdick2016dsmm,burdick2014dsmm,burdick2011deb,hernandez2010unleashing})
has lowered the barriers to such research, as has the increased availability of public 
financial documents that are typically filed with the SEC.

Latent Semantic Analysis (LSA) (\cite{LD1997}) was one of the earliest approaches developed 
to obtain semantic information from the word-document co-occurrence matrix 
of some (large) document collection. 
LSA uses dimensionality reduction based on matrix factorization and represents words and 
documents as points in a Euclidean space. 
Our research draws upon unsupervised probabilistic topic models (\cite{BNJ2003,SG2007}).
Both unsupervised topic models and their extensions have been successfully used to identify 
topics from a range of Web and social media corpora; the topics have been used to understand
user behavior and to make predictions
(\cite{Zhong2020,Chen2021,Hu2021,Liu2021,Yang2022}).
We use dynamic topic models (DTM) (\cite{BleiDTM2006}) that use time slices to 
understand the evolution of topics over time. 

Deep learning based approaches, particularly LSTM and BERT \citep{devlin2019}
are the current state of the art in document embedding. 
Recently, topic modeling approaches that use embeddings have been introduced \citep{Dieng2020}. 
However, the size of our dataset does not support such extensions;
the low document count (approximately 4,000 documents) of the \texttt{resMBS} 
dataset is insufficient for an approach such as LSTM or embedding based approaches 
that typically need vast amounts of data for training. 
Further, BERT would typically provide better models by using embeddings 
over phrases or sentences;
our dataset is limited to representing the supply chain using 
keywords and tokens for the financial institutions and their roles.

The application of these newer methods may present interesting future opportunities in 
financial research. 
We briefly summarize the results of one very successful application of 
such methods within the Business Open Knowledge Network project \cite{Pujara2021}, 
where the online Web text of company websites is obtained from 
an archival source such as the Internet Archive / Wayback Machine. 
Based on this textual information, embeddings such as Doc2Vec and RoBERTa, 
as well as topic models, were used to construct competitor networks.
These competitor networks were then used for multiple downstream 
tasks including predicting (public) companies’ profitability 
and predicting the industry sector of companies.
Preliminary results indicate that Web text and deep learning approaches can indeed play a 
successful role in understanding competition.


\section{Dataset and Estimating Financial Performance}\label{sec:dataset}

The dataset for this paper originated from the approximately 5,000 prospectuses for residential mortgage backed securities (\texttt{resMBS}) that were issued by private labels
and filed with the SEC between 2002 and 2007. 
Very few prospectuses (fewer than 50) were issued in 2008, and we exclude them from our analysis. 
We use text extraction to construct the supply chain of financial institutions 
and the roles that these institutions play in each prospectus. 
We augment the securities within a prospectus with multiple features, including
the performance history of each security, obtained from the Bloomberg repository (\cite{BloombergLP}). 
Our dataset is unique; few financial product datasets are associated with a supply chain, and even the extensive Bloomberg repository, which contains the prospectuses and the payment history, does not provide the \texttt{resMBS} supply chain. 
This knowledge is not proprietary but it is not widely available, nor is it typically used for financial analysis; this contributes to the novelty of our research. 

\subsection{Data Extraction Protocol}
We briefly discuss the protocol to extract the \emph{mentions} of financial institutions 
and the role that they played in the supply chain from the prospectuses; details are available in (\cite{burdick2016dsmm,xu2016jdiq}). The extraction pipeline is developed using the rule-based algebraic information extraction system, IBM SystemT (\cite{chiticariu2010acl}), 
and was executed within the IBM Discovery cloud infrastructure. We note that each prospectus can contain thousands of pages of semi-structured data in varying formats including ASCII text, PDF files and tokenized HTML documents. Data is often contained within tables, itemized lists, and other document elements. This leads to a typical challenge of big data volume, variety and veracity along our pipeline.

\begin{figure*}[hbtp]
\centerline{
\includegraphics[width=0.8\linewidth]{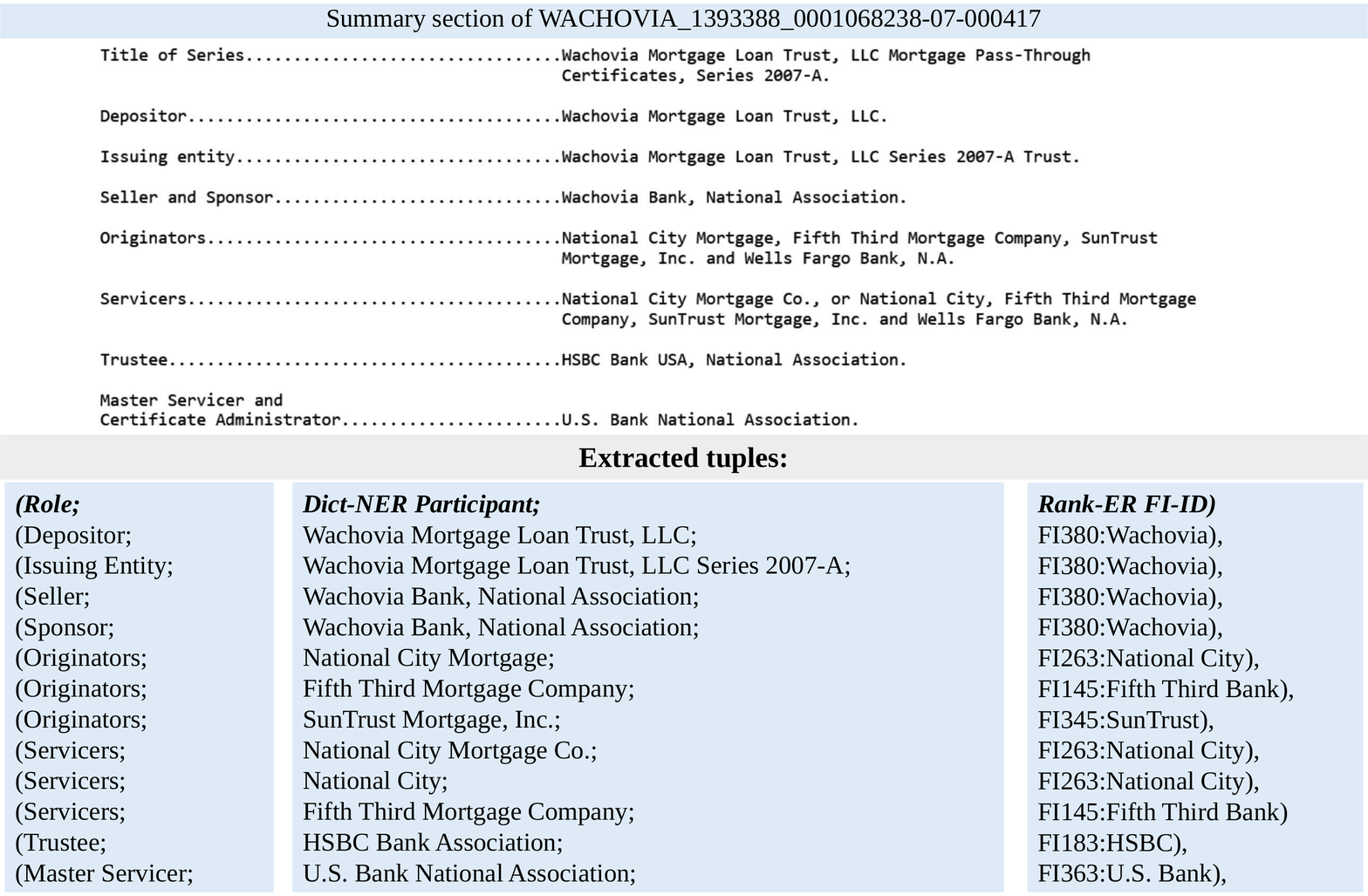}
}
\caption{Summary section of an example prospectus and the extracted 3-tuples.}
\label{fig:resMBSsummary}
\end{figure*}

We develop a Named Entity Recognizer, Dict NER (\cite{xu2016jdiq}), which is tuned to extract financial institution (FI) names from text. Financial institution names are typically composed of a root term, which is usually unique, and a suffix term, which is drawn from a small corpus of suffix terms. 
For example, \texttt{U.S. Bank National Association} is composed of the root term, 
\texttt{U.S. Bank} and the suffix term, \texttt{National Association}. 
Dict NER utilizes both a root dictionary and a suffix dictionary to recognize such names. 
We also develop an entity resolution tool, Rank ER (\cite{xu2016jdiq}), to map the extracted 
name to a corpus of standardized financial institution names. 
For example, there are multiple variants of names in Figure~\ref{fig:resMBSsummary} that are
all matched to the standard financial institution name \texttt{Wachovia}.
The standardized corpus of names was curated from multiple resources including the ABSNet portal (\cite{ABSNet}) and the National Information Center of the Federal Reserve System (\cite{NIC}). 
A Role Extraction module uses keyword matching to extract roles such as \texttt{issuer}, 
\texttt{depositor}, \texttt{originator}, and \texttt{sponsor}. 
A Role to Financial Institution Matching module pairs a role with one or more
financial institution names.

Figure~\ref{fig:resMBSsummary} illustrates the summary section of a sample resMBS prospectus. Example names in this summary include \texttt{Wachovia Bank}, \texttt{National City}, and \texttt{HSBC Bank}. 
We can also extract the role played by the FIs, e.g., depositor, issuing entity, seller, sponsor, originator, servicer, or trustee. 
As the figure demonstrates, it is necessary to correctly interpret the tabular structure in order to align the \texttt{originator} or \texttt{servicer} role with the correct set of financial institution names. 

Consider the three columns in the lower part of the figure, which contain the role, the extracted name of the mentioned entity, and the matching standardized name (determined after entity resolution). 
For this exemplar supply chain, \texttt{Wachovia} plays the role of depositor, issuing entity, seller and sponsor. 
Similarly, \texttt{National City} plays the role of originator and servicer. 
The example prospectus in Figure ~\ref{fig:resMBSsummary} will finally
be associated in the dataset with a set of pairs, e.g., 
(\texttt{Issuing entity}, \texttt{Wachovia})
and (\texttt{Originator}, \texttt{National City}). 

Details about the quality of the Dict-NER and role to FI matching are in
(\cite{burdick2016dsmm,xu2016jdiq}). 
The extraction precision of the dataset is typically between 85\% and 95\%. 
There is a wide range in precision across the documents, as
is typical for text extraction methods. 
We therefore take several additional steps to improve data quality. 
For example, we use a second extraction pipeline with a different Named Entity
Recognizer, ORG NER (\cite{chiticariu2010acl}), and only consider extracted pairs, 
associating roles with financial institution names where the two pipelines show agreement. This extension to the protocol
increases the precision of the extracted data to 90\% or higher. 

The final step of the protocol
is to obtain performance data for each security identified in each prospectus. Although each security has a unique identifier (\cite{CUSIP}), these CUSIP values are typically not generated at the time of issuance and are thus not included in a prospectus. 
The name of the issuing entity for each \texttt{resMBS} prospectus is therefore matched 
against the (issuer) names of securities available through the Bloomberg repository (\cite{BloombergLP}). 
When this automated match retrieves multiple entries, a human selects the correct matching entry. 
Finally, based on anecdotal evidence and the literature, we consider the roles of issuer and originator most important, so we filter the prospectuses to only include those where the issuer of the securities and the originator of the mortgages was successfully extracted. 

We extract multiple descriptive properties from the Bloomberg repository, including information 
on the structure of the security and tranche type, payment contingencies, and waterfall 
payoff details. For example, some securities are interest paying, or linked to floating 
interest rates, or, in the case of \texttt{senior subordinated debt (SSUP)}, payment may be subordinated to the payment of other securities 
in the prospectus.
We also obtain the MIR at issuance, as well as updated ratings. Summary statistics for the frequencies of the different roles extracted are found in Table \ref{tab:summary2} in the Appendix, and details on the attributes available from the Bloomberg repository are in Table ~\ref{tab:attributes} in the Appendix.

\begin{figure*}[hbtp]
\centerline{
\includegraphics[width=0.8\linewidth]{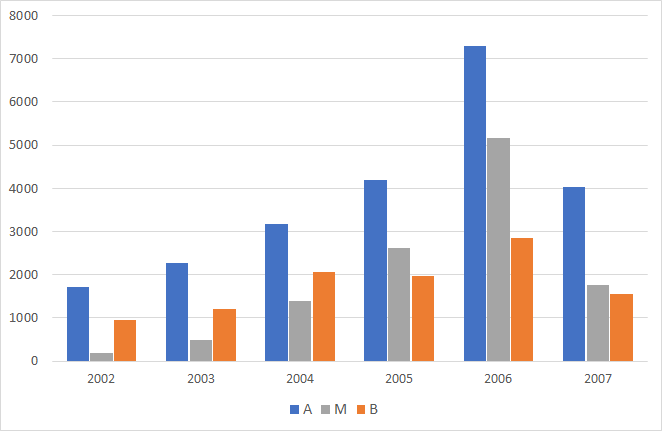}
}
\caption{The temporal evolution of the frequency of issuance of each security class, broken down by year. The summary count for the dataset is in Table \ref{tab:finperf1}}.
\label{fig:SecuritiesOverview}
\end{figure*}

Table \ref{tab:finperf1} presents summary statistics of the prospectus and security counts from the Bloomberg repository. 
These statistics are partially presented in Figure \ref{fig:SecuritiesOverview}, which highlights trends in the issuance of the 
three security classes over time. The issuance of securities grows rapidly from 2002 and peaks in 2006, with the issuance in 2007 falling below 2005 levels. 
We note that the count of \_M\_ securities increases very rapidly over this period, growing from fewer than 200 (less than 7\% of all securities) 
in 2002 to a peak of over 5,000 (approximately 34\% of all securities) in 2006. 

\begin{table}[hb]
\begin{small}
\centering
\begin{tabular} {|l| l| l| l| l| l| l| }
\hline
\hline
{} & {2002} & {2003} & {2004} & {2005} & {2006} & {2007} \\ \hline
{Count of prospectuses} & {173} & {262} & {394} & {566} & {849} & {535} \\ \hline\hline
{Count of securities} & {2,529} & {3,694} & {5,531} & {9,069} & {14,154} & {9,865} \\ \hline
{Count of \_A\_ securities} & {1,519} & {2,127} & {2,835} & {4,267} & {6,653} & {5,263} \\
{Count of \_M\_ securities} & {149} & {406} & {1,016} & {2,577} & {4,849} & {2,594} \\
{Count of \_B\_ securities} & {861} & {1,161} & {1,680} & {2,225} & {2,652} & {2,008} \\
\hline\hline
\end{tabular}
\end{small}
\caption{Summary statistics for prospectuses and securities in \texttt{resMBS} with performance data in the Bloomberg repository.}
\label{tab:finperf1}
\end{table}

\subsection{Estimating Financial Performance}

\begin{figure}[h]
\centering
\subfigure[All Securities.]
{
\includegraphics[width=7.5cm]{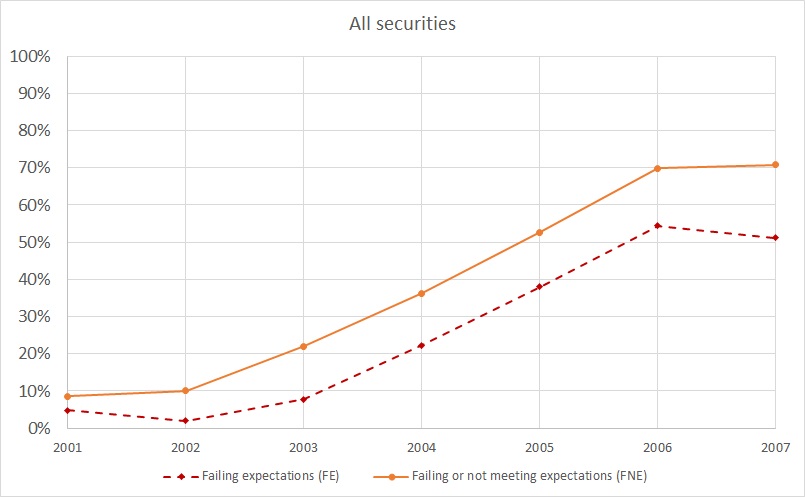}
\label{fig:FERateAll}
}
\qquad
\subfigure[\_A\_ Securities.]
{
\includegraphics[width=7.5cm]{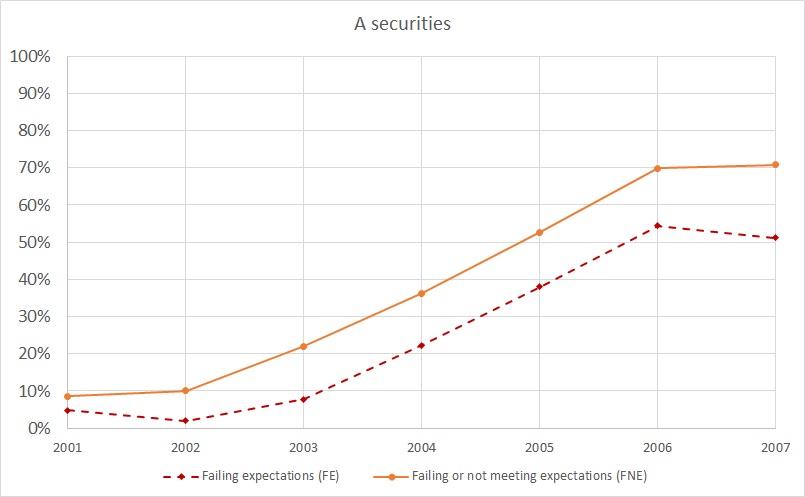}
\label{fig:FERateA}
}
\subfigure[\_M\_ Securities.]
{
\includegraphics[width=7.5cm]{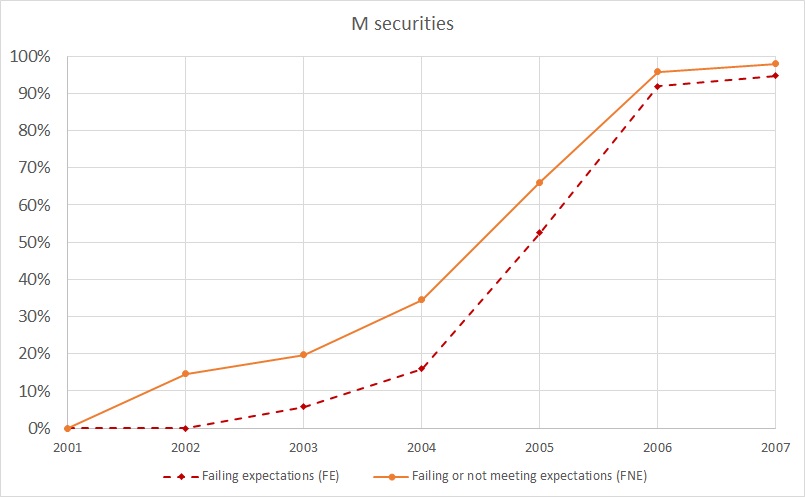}
\label{fig:FERateM}
}
\qquad
\subfigure[\_B\_ Securities.]
{
\includegraphics[width=7.5cm]{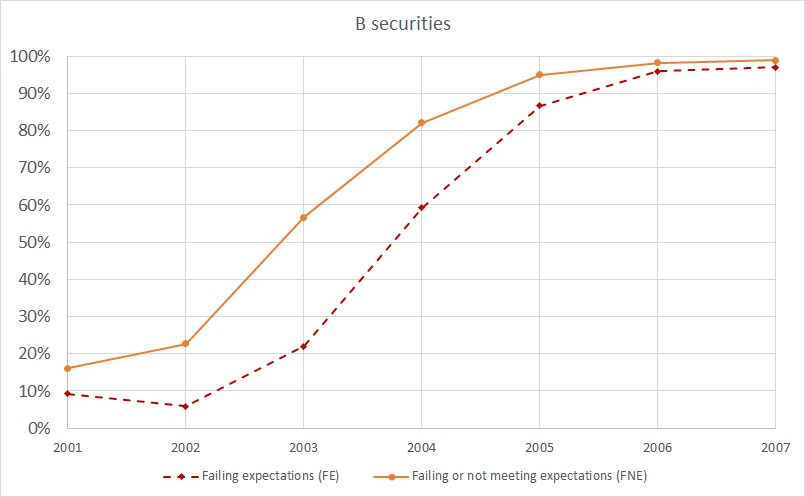}
\label{fig:FERateB}
}
\caption{
Performance trends for securities from 2002 through 2007. 
FE: failing expectations (red dashed line)
and  FNE: failing or not meeting expectations (orange solid line).
Subfigure (a) shows the trend over all securities while (b), (c) and (d)
correspond to \_A\_, \_M\_ and \_B\_ securities, respectively.}
\label{fig:failurerate_temporal}
\end{figure}

We utilize historical payment information from the Bloomberg repository to estimate 
the financial performance of each security. 
This includes information on the original mortgage amount (principal), 
the current principal balance, the sum of principal and interest payments, 
and shortfall and loss information. 
The history also indicates any early termination of payment before the principal 
payoff or maturity of the bond. 
In consultation with several investment experts, we define some (approximate) maximum 
shortfall and loss thresholds for each class, as follows:
\begin{itemize}
\item
For investment grade (Class \_A\_) securities, 
to label a security as {\em meeting expectations (ME)}, 
neither the shortfall in the principal that is repaid, nor the sum of the other shortfalls and losses,
should exceed the maximum threshold of 100 basis points (a basis point is 0.01\% of principal). 
A threshold of 2,500 basis points is the cutoff to label an \_A\_ security as 
{\em failing expectations (FE)}.
Securities whose performance lay between these two thresholds are labeled as 
{\em not meeting expectations (NME)}.
\item
For non-investment grade discounted securities (Class \_M\_ and \_B\_ securities),
we use a threshold of 500 basis points as {\em meeting expectations (ME)} 
and a threshold of 5,000 basis points as {\em failing expectations (FE)}.
Securities whose performance lay between these two thresholds are labeled as
{\em not meeting expectations (NME)}.
\end{itemize}

The models in this paper will consider two tasks. 
The first is to predict those securities that have
the label {\em failing expectations (FE)}.
The second is to predict those securities that have either the label {\em failing expectations (FE)}
or {\em not meeting expectations (NME)}; this combined group is labeled as {\em FNE}.

Figure \ref{fig:failurerate_temporal} illustrates the performance trends over time.
Figure ~\ref{fig:failurerate_temporal}(a) shows the trend over all securities, 
while (b), (c) and (d) correspond to the trends for \_A\_, \_M\_ and \_B\_ securities, respectively. 
The rates for both FE and FNE increase over time for all securities. 
For \_A\_ securities, the rate is flat from 2002 to 2004, then increases slowly after 2004. 
For \_M\_ securities, the rate starts increasing more sharply in 2003. 
In contrast, for \_B\_ securities, the rate of failure starts increasing much earlier, in 2002, 
And by 2005 these securities reach a 95\% rate of failing or not meeting expectations.


\section{Dynamic Topic Models for Financial Supply Chains} 
\label{sec:TM}
We used topic models to represent the communities that participate in the supply chain
behind the prospectuses and securities.
A topic model (\cite{BNJ2003,SG2007}) is a probabilistic model based on the idea that
a collection of documents can be described using a set of topics.
A document is modeled as a mixture over the set of topics,
with each topic represented as a probability distribution over a set of keywords in the vocabulary of that collection.
In our case, the keywords are the pairs of financial institutions and their roles.
However, the financial institutions, roles, and communities were not static
over our time horizon. 
Some institutions left a community or played a different role, while other financial 
institutions may have joined the community.
To capture this evolution of topics over time, we adopted time series extensions to the topic models that have been introduced.
This includes a continuous non-Markovian model of topics over time (ToT) (\cite{Wang2006})
and dynamic topic models (DTM) (\cite{BleiDTM2006}).
We used the DTM extension, in which the document
collection is also divided by time – in our case, one time slice for each year. 
The Appendix briefly introduces topic models and DTM
(\cite{BNJ2003}).

\subsection{DTM Experimental Results}

\noindent{\bf Configuration}

We used the Python $gensim$ implementation of DTM (\cite{rehureklrec}), which represents each document (prospectus) as a bag of words over a vocabulary of
(\texttt{Role}, Financial Institution) pairs.
We filtered the data and eliminated financial institutions that appeared in fewer than 20 prospectuses, as well as prospectuses that had fewer than five pairs.
This created a subset of 4,472 prospectuses.
Additional statistical information on these prospectuses is provided in
Tables \ref{tab:summary1} and \ref{tab:summary4} in the Appendix.

Some roles such as \texttt{issuer} and \texttt{originator} are particularly important in the supply chain.
Others such as \texttt{servicer} may occur very frequently but 
do not play a central role in creating prospectuses nor significantly impact financial performance. To address this variation in role importance,
we doubled the token weight associated with pairs that involve the roles of
\texttt{issuer} or \texttt{originator}.

We created annual time slices from 2002 to 2007 using the date of the prospectus issuance.
(We note that the data is insufficient for us to consider both month and year.)
We varied the hyperparameter $\alpha$, which affects sparsity of the topics in each time slice; however, this parameter did not have much impact. 
We also varied the count of topics from 20 to 50 in increments of five. We did not observe
much change in the community structure beyond 30 topics,
so we report results for 30 topics.

The hyperparameter $top\_chain\_var$ is significant since it controls the rate at which the topic evolves with each new time slice. 
Setting the parameter at one extreme will convert DTM to a static topic model,
while setting it at the other extreme will result in independent topics in each time slice.
We used two settings for this parameter, 0.005 and 0.75.
A low value of 0.005 results in slow evolution while a high value of 0.75 results in fast evolution.
We labeled the results with $top\_chain\_var$ = 0.005 as $DTM_{slow}$ and $top\_chain\_var$ = 0.75 as $DTM_{fast}$. To set a baseline for topic evolution, we computed the static LDA topics, labeled as $LDA$.

Completeness is an intrinsic problem for unsupervised methods such as topic models, 
where there often is no independent ground truth data. 
Human evaluation is typically used as a basis for evaluating both the accuracy and completeness of topics over 
document corpora. 
Here, we performed an indirect evaluation of accuracy and completeness as follows.
First, we noted that a large number of prospectuses are strongly associated with a single topic;
this indicates that the topics capture strong co-occurrence patterns in the data. 
For example, 71\% of the prospectuses/securities in the dataset are associated 
with a single topic, with a topic weight of at least 0.7. 
Second, our models demonstrated that topics/communities are found to be 
significant when predicting financial performance. 
Together, these observations imply the following:
the topics are strongly associated with a significant fragment of the dataset;
the topics provide good coverage of the securities and prospectuses; and
the topics are associated with financial performance.
This then suggests that there is no evidence of missing topics 
and that the topics are reasonably accurate and complete. 

\noindent{\bf DTM Result Summary}

The $DTM_{slow}$ model showed minimal temporal evolution, and the topics were very similar to $LDA$ (by experiment design).
We therefore compare the topics in the $DTM_{fast}$ model with the (union of) topics in $LDA$ and $DTM_{slow}$. We observe three types of communities:

\begin{itemize}
\item
\texttt{Stable:} Twelve topics that appear in $DTM_{fast}$ are conserved and appear to
be well aligned to topics in both $LDA$ and $DTM_{slow}$.
There is very low evolution of topic composition across the time slices for these stable topics. In other words, these topics represent communities in the supply chain that do not
evolve over the time period. 
\item
\texttt{Evolving:}
These are topics that appear in $DTM_{fast}$ and show some overlap with topics in
$LDA$ and $DTM_{slow}$; however, the topics are not fully aligned.
In several \texttt{evolving} $DTM_{fast}$ topics, we observe an evolution of the
\texttt{issuer} associated with the topic.
\item
\texttt{Dynamic:}
Of most interest are eight topics that only emerge in $DTM_{fast}$.
In several cases, these topics may be associated with smaller financial institutions that
issued a moderate volume of prospectuses.
Consequently, these topics may have been merged with some other topics in the
$LDA$ or $DTM_{slow}$ models.
The additional modeling flexibility of DTM, which explicitly models the community structure
for each time slice, facilitates their appearance.
Hence, the emergence of \texttt{dynamic} topics clearly highlights the benefits of using DTM.
\item
Seven small topics (with fewer than 25 prospectuses each) were not studied further.
\end{itemize}

\subsection{Exemplar Topics}
\subsubsection{The supply chain}

We note that some evolving or dynamic communities may be focused on the issuer,
while the originators of mortgages or service providers may join or leave a community over time.
Alternatively, some communities may focus on the originators of the mortgages;
as the community evolves, different issuers may engage with these originators
and their loans may be pooled within prospectuses from these different issuers.
In some cases, there may be a continuous evolution of financial institutions across the roles.

\begin{figure}[h]
\centering
\subfigure[Topic 8]
{
\includegraphics[width=7.5cm]{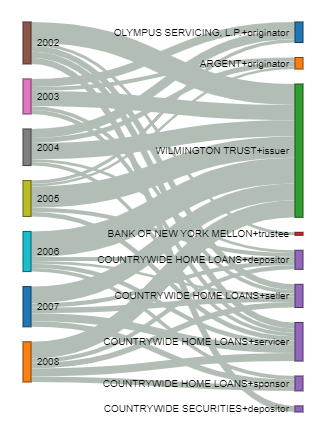}
\label{fig:topic8}
}
\qquad
\subfigure[Topic 26]
{
\includegraphics[width=7.5cm]{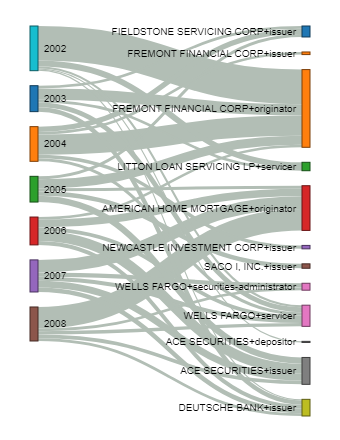}
\label{fig:topic26}
}
\caption{Two exemplar supply chains, corresponding to Topic 8 (a)
and Topic 26 (b). The nodes on the left represent the community for that year and link to 
the (Role, Financial Institution) pairs on the right; the thickness of the edge represents the strength of the connection. }
\label{fig:sankey}
\end{figure}

Figure \ref{fig:sankey} summarizes two topics (Topics 8 and 26) with different time dynamics.
The illustration is a Sankey diagram rendering of a bipartite graph.
Each of the nodes on the left represents the community for that year.
The nodes on the right are the (Role, Financial Institution) pairs.
The thickness of the edge represents the weight or significance
of the financial institution and role to the topic.
The supply chain represented by Topic 8 in Figure \ref{fig:sankey}(a) is
\texttt{stable} over the time slices.
\texttt{Wilmington Trust} is the issuer and \texttt{Countrywide} participates in many roles, including servicer, seller, depositor and sponsor, across all time slices.

The supply chain represented by Topic 26 in Figure \ref{fig:sankey}(b) is \texttt{dynamic}.
\texttt{Fremont} served as the originator in the early time slices, but that role is then
assumed by \texttt{American Home Mortgage} starting in 2005.
There is also an evolution of issuers.
\texttt{Fremont} and \texttt{Fieldstone} are issuers in the early time slices, and
\texttt{Ace Securities} and \texttt{Deutsche Bank} join as issuers in later time slices.

\subsubsection{Communities and financial products }

We highlight four communities. As previously discussed, Topic 8 is a large \texttt{stable} topic while Topic 26 is \texttt{dynamic}. 
We also consider Topics 11 and 28. 
Topic 11 is a \texttt{stable} topic where \texttt{Bank of America} is active in multiple roles
through the different time slices. 
Topic 28 is \texttt{dynamic}, with \texttt{Renaissance} and \texttt{Principal Residential} as issuers.
These four communities are each associated with 100 to 200 prospectuses and an average of
approximately 1,500 securities. 
The largest (Topic 8) is associated with approximately 200 prospectuses and more than 3,000 securities.

\begin{figure}[h]
\centering
\subfigure[Activity per year]
{
\includegraphics[width=7.5cm]{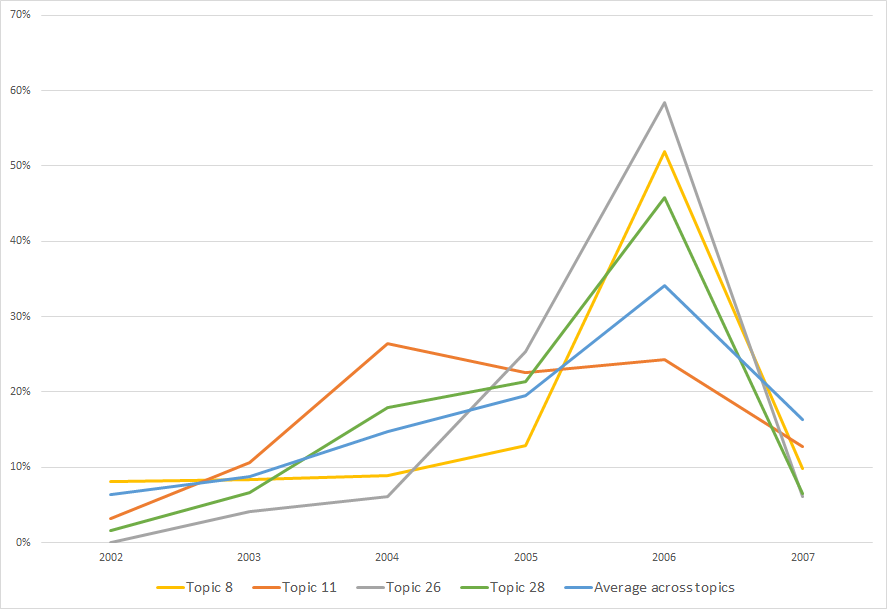}
\label{fig:topicvolbyyear}
}
\qquad
\subfigure[Composition by class]
{
\includegraphics[width=7.5cm]{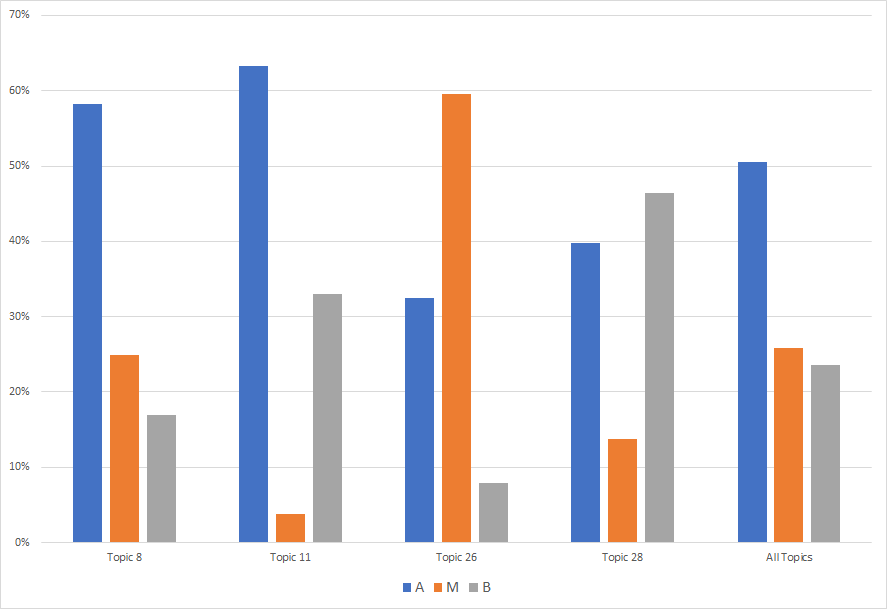}
\label{fig:classbytopic}
}
\caption{Heterogeneous characteristics of Topics 8, 11, 26 and 28. 
Subfigure (a) shows the normalized fraction of securities issued per year 
for each topic, contrasted with the sample average. 
Subfigure (b) shows the normalized fraction of securities by class, again contrasted with the sample average.}
\label{fig:topic_dd}
\end{figure}

These communities represent a range of behavior. 
Their activity level related to issuing prospectuses varied across the time slices, and the composition of the prospectuses (by security class) varied across topics.
Figure \ref{fig:topic_dd} (a) highlights the peak activity of each topic, based on the
normalized fraction of prospectuses issued per year (normalized based on the overall 
count of prospectuses for that topic). 
We note that Topic 11 peaks early in 2004, while the other three topics peak in 2006. 

The varied composition of the prospectuses is even more interesting.
Figure \ref{fig:topic_dd} (b) summarizes the composition by security class for each topic.
The overall composition across all topics is shown at the far right of the figure.
When contrasting the securities composition of each topic, we observe that Topic 8 
most resembles the overall composition across all topics.
Topic 11 produces very few Class \_ M\_ securities, while Topic 26 produces an excess.
Almost 60\% of Topic 26 securities are Class \_ M\_ securities; the sample average is approximately 25\%. 
Topic 28 produces an excess of approximately 50\% Class \_B\_ securities; 
the sample average is just over 20\%. 
Clearly, the types of prospectuses that are produced vary from one community/topic to the next. 

\begin{figure}[hbtp]
\centerline{
\includegraphics[width=0.6\linewidth]{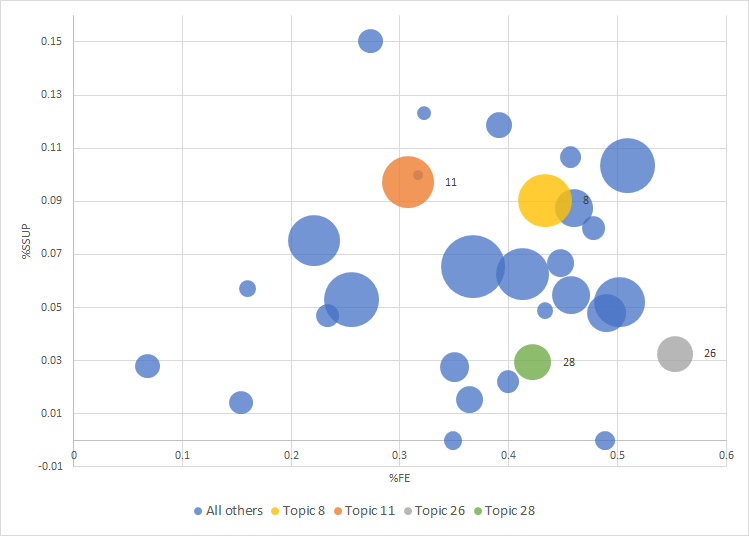}
}
\begin{center}
\caption{Performance summary of securities in each community. The x-axis represents the percentage of securities that fail expectations (FE), and the y-axis is the fraction of securities labeled \texttt{SSUP}. }
\label{fig:topic_rawperformance}
\end{center}
\end{figure}

Figure \ref{fig:topic_rawperformance} summarizes the overall financial performance of each topic. 
The x-axis represents the percentage of securities that fail expectations (FE), and the y-axis represents 
the percentage labeled \texttt{SSUP} (senior subordinated debt, which as we will highlight in the next section, is a key indicator of financial performance). 
The FE percentage across topics ranges from the single digits to over 50\%,
and the percentage of securities that are \texttt{SSUP} ranges from 0\% to 15\%.

For our focal topics, we note that Topics 8 and 11 both issue a large fraction of 
\texttt{SSUP} securities but differ significantly in their FE rate. 
Topic 8 has a high FE failure percentage of 47\%, while
Topic 11 has a moderately high percentage of 33\%.
In contrast, Topics 26 and 28 issue a smaller fraction of \_A\_ securities and
\texttt{SSUP} securities.
Recall that \texttt{SSUP} is a label for \_A\_ securities, 
so a low \_A\_ issuance necessarily leads to low \texttt{SSUP} issuance. 
We note, however, that the failure rate of Topic 26 is over 50\%, 
the highest of all the topics. 
This high failure rate can in part be explained by the high proportion 
of \_M\_ securities issued.
The failure rate for Topic 28 is lower, at just over 40\%.

We further emphasize that the differing prospectus composition
and performance range across the four communities
is notable since we do not use any of this information in the DTM model. 
The question remains whether the topics are significant in predicting financial performance, and we explore this question in the following section.


\section{Models to Predict Performance}\label{sec:pred_models}

In this section, we present a sequence of increasingly comprehensive models 
to help us understand the factors that impact the financial performance of \texttt{resMBS} securities.
We start with a model that includes the features of individual securities. 
Next, we consider a model that further includes prospectus level features. 
The final model includes supply chain features, represented by the topic weights.

As previously noted, we build our models for two binary outcomes: 
(i) securities that fail expectations (FE) and 
(ii) securities that either fail or do not meet expectations (FNE).
Details and statistics related to these outcome labels were provided in Section 3.2 and Figure 5.
We report on the results of a regularized logistic regression LASSO model (\cite{TIB1996}). 
The LASSO approach benefits from ease of regularization and the interpretability of the resulting models. 
It also compared favorably with other machine learning approaches ranging from 
simple classification trees to ensemble approaches for this dataset and task(s). 
For each model, we report on two metrics: accuracy and the F1 score.
The F1 score is defined as the harmonic mean of precision (positive predictive value) and 
recall (sensitivity) of the positive labels.

\subsection{Model Design and Potential Limitations}

We first discuss the motivation for our modeling approach and then its limitations.
In a context like ours, where potentially hundreds of features must be considered, 
feature engineering must be carefully considered. 
The information about the supply chain that is extracted from prospectuses
can be represented in a number of different ways. 
For example, we could use a binary indicator that directly reflects whether a financial 
institution participated in the supply chain for a particular security. 
Alternatively, we could use a binary indicator to represent a specific 
financial institution playing a specific role, e.g., servicer. 
However, we chose to represent the supply chain with a topic weight feature 
serving as a proxy for the community, 
with the community comprising multiple financial institutions and their roles. 
This topic weight feature can be viewed as a summary statistic for the supply chain, and 
it has the added benefit of being more dense than the individual binary indicators.

Another key concern when choosing a model is the ease of interpreting results.
The major motivation for selecting our three increasingly comprehensive models is 
that these models accurately reflect our understanding of how the features of the securities,
the composition and complex waterfall payment structure of the prospectuses, 
and finally the financial supply chain, all impact performance. 
The relationships among features at the different levels may be complementary and/or correlated. 
These three models assist in our increasingly nuanced understanding and explanation of the 
financial insights that we obtain from the dataset.

First, each \texttt{resMBS} security is characterized by a set of features that reflects
the security class within the waterfall structure, the rating and the year of issuance.
In the absence of other factors, these features are the drivers of financial performance, and
we indeed find these security level features to be significant in the security
level model. 

Our summary statistics show that the distribution of securities across
the three security classes, \_A\_, \_M\_ and \_B\_, varies across each of the prospectuses. 
Further, there are temporal trends in this variation, and these trends appear to
impact performance. After consultation with experts, 
we constructed a set of features that captures these prospectus level variations. 
For example, within a prospectus, the presence of even a single security with the label 
\texttt{Senior Subordinated debt (SSUP)} could impact the 
financial outcome for all other securities in that prospectus, 
independent of whether these other securities are labeled \texttt{SSUP}. 
Thus, the presence of an \texttt{SSUP} security is an overall quality signal for a prospectus. 
It was through a deep dive into the waterfall model of the payouts, and extensive consultation with experts,
that we were able to identify 
these important prospectus level features that impact performance.

In this context, it is noteworthy that we experimented with a hurdle model. 
First, we built a model to predict the overall performance of the prospectus, 
followed by a model of the performance of the individual securities. 
This more sophisticated model performed well, but it did not lead to significant 
improvements in our performance measures; 
thus, we could not justify the increased modeling complexity. 

Next, we addressed our representation of the supply chain. 
As discussed earlier, when we constructed our security and prospectus level models, 
we considered categorical features reflecting the participation 
of a financial institution. 
However, with hundreds of financial institutions playing a number of different roles, 
the data was too sparse to yield meaningful results. 
Thus, the direct inclusion of the financial institution as a feature 
(or financial institution playing a specific role)
was not helpful. 
We also experimented with representing the key players, e.g., issuers, 
and their level of involvement in the supply chain, 
e.g., the count of different roles played by one issuer. 
We determined that the topic weight was the best feature to be a proxy for the communities, 
and it was able to provide real-world financial insights. 

The key motivation for our use of the three models is that each of these models focuses 
on a set of financial features (security or prospectus or topic) that represents 
some aspects of the dataset. 
By providing three separate models, we can connect each model and its outcomes to 
specific financial insights. 
This allows us to provide more detailed explanations.
For example, we can highlight the fact that many toxic and partially toxic topics are 
associated with an increased risk of failure, 
even after accounting for security and prospectus level features. 

We note that as we move from the security to the prospectus level models,
the significant features and their coefficients from the security level model appear
to be conserved in the prospectus level model. The features at these two levels
appear to be complementary in their impact on financial performance.
However, this is not the case when we move to the third comprehensive model;
there is a change in the significant features and their coefficients.
This suggests that some communities or topics in the supply chain 
may be correlated with the underlying prospectus and security features. 
We use the three models, and this potential correlation, to provide more nuanced insights 
for some partially toxic topics.

In addition to the models presented in this section, we developed more detailed models. 
For example, we built independent models for each security class and year of issuance. 
Additionally, we incorporated time varying security features, hypothesizing that the 
impact of features such as \texttt{SSUP} were not constant over the study horizon. 
Although these additional models were comparable in performance 
(as measured by accuracy and F1 score), 
they did not provide additional insights, and we could not justify 
the additional modeling complexity. 

The models were implemented using the \emph{glmnet} package in R. 
Tuning was done using 10-fold cross validation and a mean square error loss. 
When splitting the data across folds, we ensured that we did not train on securities 
from a particular prospectus, which would lead to subsequent testing on securities from the same prospectus. 
We note that the search range for the regularization parameter had to be changed significantly from the default values.

Due to the shrinkage of the LASSO model, the model is not unbiased, and the regression coefficients 
cannot be directly interpreted as the log odds of the outcome. 
Almost all the features are binary or have values between zero and one; 
the only exception is the original mortgage amount. 
Thus, despite the estimation bias, the magnitude of the resulting regression coefficient 
can be an approximate indicator of the relative importance of that feature in 
predicting the outcome. 
We note that many features may be correlated, and this may contribute to the presence 
and magnitude of the regression coefficients.

Finally, we carefully considered the many limitations of our retrospective data, 
including the potential for endogeneity. 
This included investigating the time varying effects of all of the 
independent variables, as discussed above. 
Yet in considering how we would test for potential endogeneity issues, we noted that there is no obvious 
choice of an instrumental variable for the resMBS dataset. 
Most financial research uses a portfolio of assets (stocks or bonds) to create a 
treatment dataset and a control dataset, but such a portfolio approach is not feasible here, as 
we cannot directly manipulate the resMBS dataset. 

However, we also recognized that the year 2007 is critical in the timeline of the financial crisis, 
and this feature value 
may be a proxy for many unobserved real-world factors that affect the securities’ performance. 
We therefore include an annual control variable, the year of issue for the securities, 
to capture, for example, the time varying effect from the economy. 
Without these annual controls, financial communities operating in 2007 would appear {\em worse} 
than communities that were active in, say, 2002. 
In other words, in order to observe whether financial communities had any impact 
beyond their operating environment and the product characteristics, 
we included the annual controls as well as all the other key product features. 
Due to the use of this annual control variable, we could not use a traditional time 
dependent training / testing split, such that 
the training data for each year precedes the testing data.
For example, there was insufficient data to use, for instance, 
the month of issuance, to create the train / test split.

We note that endogeneity can also result when some treatment is not applied 
randomly and when there is selection bias in dataset sampling. 
In this study, we considered all of the prospectuses issued in the period of 
the experiment, with the caveat that there was a limited number of prospectuses 
from which we were unable to extract data or which we otherwise filtered out because they had insufficient data. 
Thus, we do not introduce a selection bias that may lead to additional endogeneity concerns.

\subsection{Insights from the Security Level Model}

Features of this model include the security class, the year of issuance, 
the MIR and several characteristics of the waterfall payoff structure.
There are over 60 values for the MIR. We aggregated these values to a higher level rating,
where \emph{Aaa} represents the securities that have the lowest credit risk. 
The \emph{Aaa} rating is followed, in increasing order of risk, 
by \emph{Aa}, \emph{A}, \emph{Baa}, \emph{Ba}, \emph{B}, \emph{Caa}, \emph{Ca} and \emph{C}. 
Some securities were evaluated but did not have a rating; 
we label this value \emph{MIR\_NR}. 
Finally, securities that were not evaluated for an MIR value are labeled \emph{MIR\_null}. 

The waterfall payoff characteristics for the security class or tranche are captured 
using 73 binary variables. 
As an example, the following two features are key to identifying the placement of the 
security in the waterfall payoff structure shown in Figure \ref{fig:rmbs_waterfall}: 
\texttt{Super Senior Bond (SSNR)} indicates that the principal and/or interest 
distributions for this security have priority over other senior securities, 
while \texttt{Senior Subordinated debt (SSUP)} indicates that the
principal and/or interest distributions are subordinated to other senior bonds
within the prospectus.

\begin{figure}[h]
\centering
\includegraphics[width=0.9\textwidth]{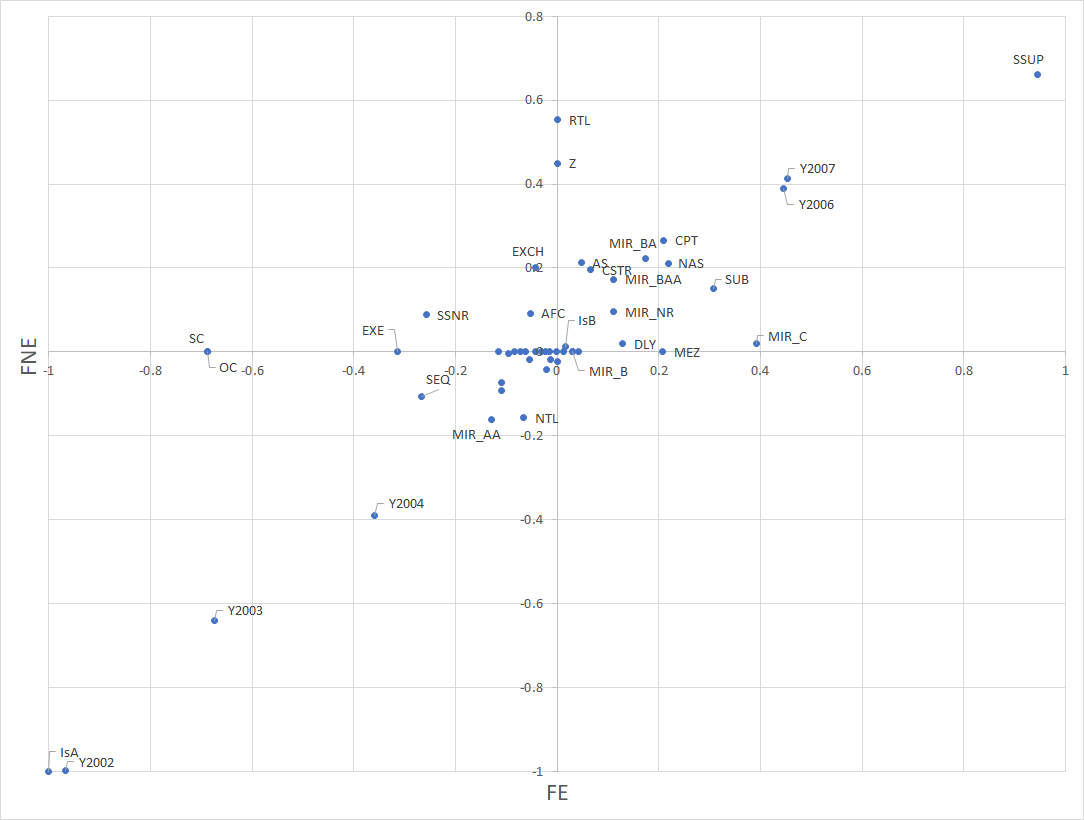}
\caption{Key factors for the security level models for the two outcomes FE and FNE.
The regression coefficients are normalized such that the magnitude of the largest coefficient is 1 for each model.
\emph{IsA} is an indicator variable for class \_A\_ securities. 
The regression coefficients are found in Table \ref{tab:model1coeff} in the Appendix.}
\label{fig:securities_FE_NME}
\end{figure}

Figure \ref{fig:securities_FE_NME} summarizes the results of the security level model
for the two outcome variables, FE and FNE. 
The x-axis has the values of the coefficients for FE, and the y-axis has the values of 
the coefficients for FNE. 
The accuracy of the FE and FNE models are 91.3\% and 84.7\%, and the F1 scores are 
0.891 and 0.857, respectively. 

A significant effect is that the annual controls for both 2007 and 2006 increase the 
predicted probability of FE and FNE. 
Meanwhile, the annual controls for 2002 and 2003 and the \_A\_ security class decrease the 
predicted probability of FE and FNE. 
This is consistent with the trends observed in Figure \ref{fig:failurerate_temporal}. 

The MIR value of \emph{MIR\_NR} increases the probability of both FE and FNE.
The MIR values of \emph{Baa}, \emph{Ba}, and \emph{C} all increase the predicted 
probability of FE, while a value of \emph{Aa} decreases the same probability. 
The MIR value \emph{Aa} was primarily assigned to \_M\_ securities. 
Over 75\% of \_M\_ securities have this rating, and they are on average less risky than \_M\_ securities in general.

After accounting for the security class and the year of issuance, many of the waterfall 
payoff features associated with a security class or tranche were retained. 
Notably, the presence of the label \texttt{Senior Subordinated debt (SSUP)} 
significantly increases the probability of both FE and FNE. 
The impact of \texttt{SSUP} on performance is consistent with the waterfall structure.
Several other waterfall payoff features also increase the predicted probability of 
both FE and FNE. 
This includes \texttt{CPT}, which is an indicator that each component of the payback 
may vary, and \texttt{sequential pay (SEQ)}, an indicator that the principal will 
be paid in sequence, after the principal of higher-priority securities reaches zero. 

Some features decrease the predicted probability of FE but do not impact FNE. 
This includes \texttt{Over Collateralization (OC)} and \texttt{Excess (EXE)}
-- both indicators that these securities have the rights to certain excess interest 
and principal payments – and \texttt{SC}, which indicates securities that are 
backed by structured collateral.
These features reflect some \emph{middle ground} since they decrease the probability of FE, 
but they also do not guard against the shortfalls and payoff losses of FNE. 
Finally, some features have no impact on FE, but they increase the probability of FNE. 
These features include \texttt{Retail (RTL)}, an indicator that the security is 
designated for sale to retail investors, and \texttt{Accrual (Z)}, which indicates 
that the accruing interest is added to the outstanding principal balance (for some period). 

We note that the impact of the label \texttt{Senior Subordinated debt (SSUP)} is the 
largest of any individual feature, after accounting for the security class and 
the year of issuance. 
In addition, several waterfall payoff features also have a major impact. 
This may indicate that given the complexity of the \texttt{resMBS} waterfall 
payoff structure, the MIR values alone cannot fully represent the nuanced risk. 

To summarize, the security level model is consistent with the observed trends of 
Figure \ref{fig:failurerate_temporal} and provides valuable insight into the 
waterfall payoff structure. 
The model is accurate in its predictions for individual securities.
However, we are interested in gaining more insights into 
the drivers of poor financial performance for \texttt{resMBS} securities.
To do this, we next introduce prospectus and supply chain level features in our models.

\subsection{Prospectus Level Model}

We hypothesized that prospectus level features may provide additional insight into financial performance. We therefore used 13 variables to capture the composition of a prospectus. These variables include the count and fraction of securities and the nominal volume and fraction of the volume (in USD) in each of the three security classes. 
In addition, given the significance of the feature \texttt{Senior Subordinated debt (SSUP)}, 
we further hypothesized that the presence of \texttt{SSUP} securities within a prospectus 
may be significant; we used a binary indicator (HasSSUP) to signal when a 
prospectus contains at least one \texttt{SSUP} security.

\begin{figure}[h]
\centering
\includegraphics[width=0.9\textwidth]{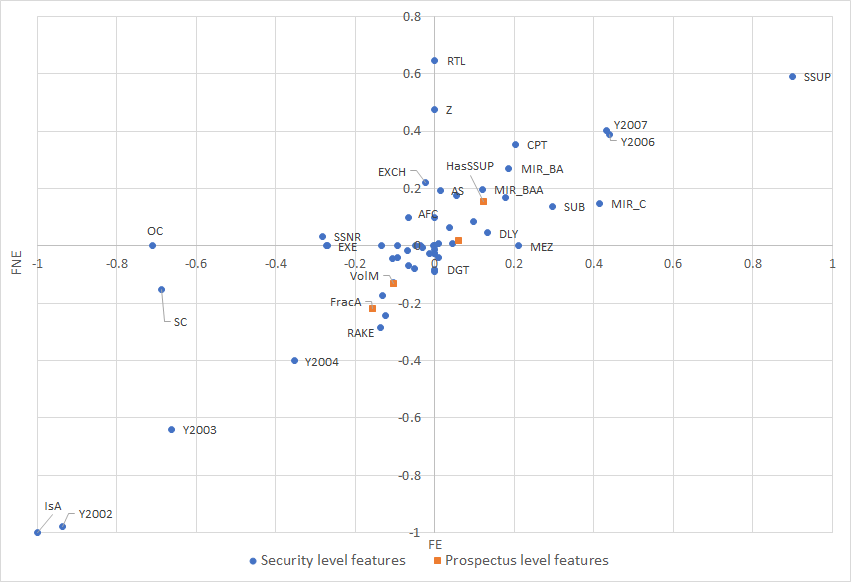}
\caption{Key factors for the prospectus level models for the two outcomes FE and FNE.
The regression coefficients are normalized such that the magnitude of the largest coefficient is 1 for each model.
The regression coefficients are found in Table \ref{tab:model2coeff} in the Appendix.}
\label{fig:securities_pros_FE_NME}
\end{figure}

The accuracy and F1 scores for the prospectus level model are similar to those for the security level model. Figure \ref{fig:securities_pros_FE_NME} provides the normalized coefficients. 
Many of the observations from the security level models hold:
the security class, the year of issuance, MIR values, some waterfall payoff features and 
\texttt{SSUP} continue to have a strong impact on both FE and FNE.

We observe that new to this model, the fraction of securities in class \_A\_ decrease the predicted probability of FE and FNE, which is to a smaller extend counter balance by the impact of the nominal value (in USD) in these same securities.  The nominal volume (in USD) in \_M\_ securities decreases these probabilities. Most notably, the indicator HasSSUP increases the probability of both FE and FNE. This is because if the model indicates the presence of even one security with the label \texttt{SSUP} within a prospectus, it can increase the probability of both FE and FNE for all other securities in that prospectus, independent of whether these other securities are labeled \texttt{SSUP}.

In summary, the model reflects the significance of prospectus composition 
in predicting the financial performance of the securities. 
Prospectuses that include \texttt{SSUP} securities 
and prospectuses with an excess of class \_M\_ and \_B\_ securities
increase the predicted probability of FE and FNE for {\em all securities} within the prospectus. 
We note that the impact of the fraction of class \_M\_securities is counterbalanced by their volume. 
This is a notable finding, as one may reason that the excess of class \_B\_ (or \_M\_) 
securities may serve as a buffer to reduce the risk for the senior 
class \_A\_ securities. Our model and our observations show the opposite effect:
an excess of class \_B\_ (or \_M\_) securities is a negative signal that 
increases the risk for senior class \_A\_ securities.

We further hypothesize that the prospectus may be a surrogate for the supply chain
that produced the securities. This suggests that some supply chains may have produced prospectuses with varying composition across the security classes and correspondingly varying risk profiles. We study the supply chain in the next model.

\subsection{Comprehensive Model}

\begin{figure}[h]
\centering
\includegraphics[width=0.9\textwidth]{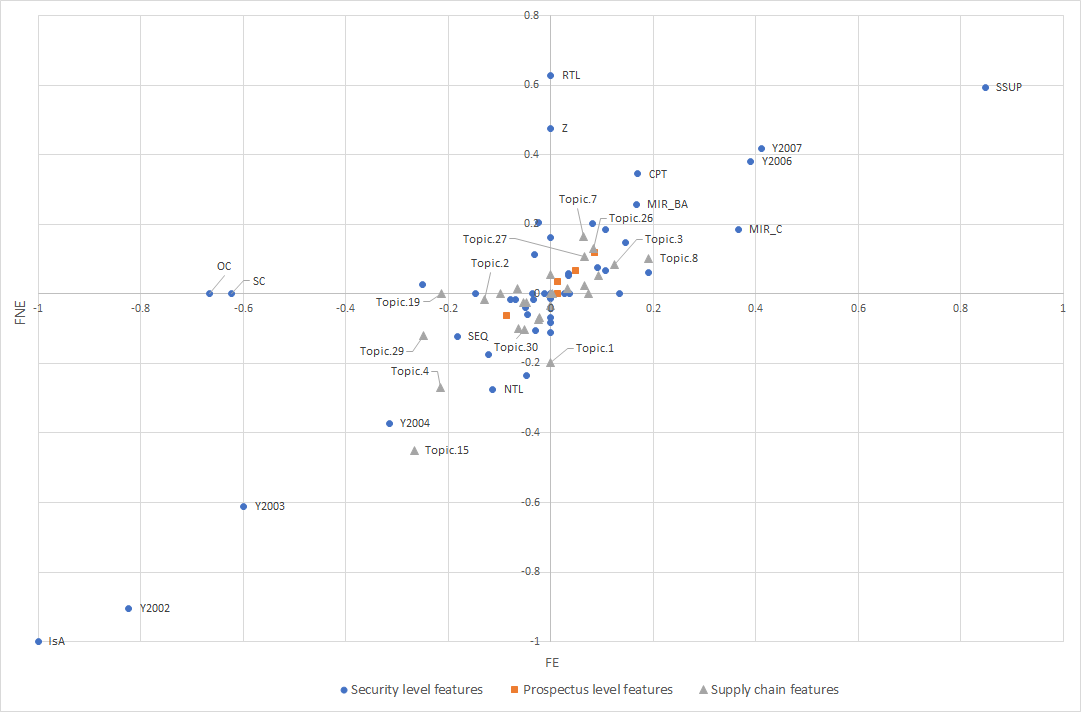}
\caption{Key factors identified as predictive by the LASSO models for the two outcomes. 
The regression coefficients are normalized such that the magnitude of the largest coefficient is 1 for each model. 
The full regression coefficients are found in Table \ref{tab:model3coeff} in the Appendix.}
\label{fig:securities_pros_topic_FE_NME}
\end{figure}

Our final comprehensive model includes all features of the Security and Prospectus
models. In addition, we used 30 indicator variables, one for each topic.
The DTM model assigns a weight vector over the 30 topics to each prospectus.
We selected the largest topic weight for each prospectus and its securities. The coefficients of this model are summarized in
Figure \ref{fig:securities_pros_topic_FE_NME}. 
The x-axis shows the values of the coefficients for FE and the 
y-axis shows the values of the coefficients for FNE.

Of the 30 binary indicators for the communities, 23 were retained by the FE model 
and 24 were retained by the FNE model. 
The coefficients for the topics have similar weights 
in comparison to features in the Security Level and Prospectus Level models.
In other words, after accounting for the security and prospectus features, 
the supply chain has \emph{additional significant impact} on the 
predicted performance. 
We discuss the financial insights to be obtained from the impact of the 
topics on performance in the next section.

\clearpage
\section{Financial Insight from the Topics in the Comprehensive Model}

Recall that our strategic goal is to use the topics (communities) to obtain financial insights
on the performance of subsets of prospectuses and securities.
In this section we take a deep dive into the financial impact of communities. 
Using a variety of real-world evidence (e.g., Bloomberg company profiles, news articles, 
financial press articles, Wikipedia), we identify institutions that were active in the subprime 
market and failed or faced penalties during or after the crisis. 
We label these institutions as toxic. A community that is closely associated with one or more 
such toxic institutions is labeled as toxic or partially toxic. 
We next review the financial impact of the topics based on their significance in our comprehensive model. 
We demonstrate that the toxic topics and some partially toxic topics increase the risk of failure,
while all of the non-toxic topics decrease the risk of failure. 
Some partially toxic topics are not significant in the comprehensive model and display more nuanced behavior; 
we will explore them in more detail in this section.

We conducted a comprehensive review of the prominent financial institutions for each topic. 
We excluded (six) topics that were small (fewer than 40 prospectuses). 
We further excluded (two) evolving topics where no financial institution appeared to play a prominent role; 
these topics appeared to reflect a high turnover of financial institutions and the
presence of multiple communities. 
We use the following protocol to label the remaining twenty two topics, where we identify
the following factors as contributing towards toxicity:
\begin{itemize}
\item
An institution filed for bankruptcy or was subject to regulatory fines.
\item
An institution underwent an involuntary merger.
\item
An institution received federal bailout funds from the Troubled Asset Relief Program (TARP).
\item
An institution was closely associated with the
subprime market and encountered financial difficulty, but may have survived the financial crisis.
\end{itemize}
An institution was labeled as toxic if it satisfied either of the first two criteria. 
The third and fourth criteria were used to label an institution as partially toxic. 

The protocol for labeling communities based on the toxic or partially
toxic labels for the financial institutions is as follows: 
A community is considered toxic if there was a significant presence of toxic institutions 
in the community in the key roles of issuer or originator. 
We used a threshold that at least two different toxic institutions participated 
in the community over the time interval, or one toxic institution played many roles over multiple years. 
A community is labeled non-toxic if it was not associated with any toxic institutions. 
A community is labeled partially toxic if it was neither toxic nor non-toxic.
For example, a community associated with a single toxic financial institution playing a 
single role may be labeled as partially toxic. 

The results of this labeling is as follows:
eight topics had no association with toxic financial 
institutions and were labeled as non-toxic 
(topics 2, 4, 5, 6, 16, 17, 28, 30). 
Of the remaining fourteen topics, five are labeled as toxic (associated with multiple toxic institutions) 
while the remaining nine are partially toxic.

To confirm our manual labeling, we constructed a (simpler) model that
only considers the security class, the topic, and the annual control as independent variables; 
i.e., the model did not use the additional security and prospectus level features that 
were included in the comprehensive model. 
This simpler model confirmed that the non-toxic topics all either decreased the risk of failure,
or had no impact on performance, for both FE and FNE. 
Similarly, the toxic or partially toxic topics all either increased the risk of failure, 
or had no impact on performance, for both FE and FNE. 
\emph{Further, all eight non-toxic topics were found to decrease the risk of failure, 
for both FE and FNE, in the comprehensive model.}

Next, we took a deep dive into the fourteen toxic and partially toxic topics; 
they are summarized in Tables \ref{tab:mixtopics1} and \ref{tab:mixtopics2}.
The summary includes the count of prospectuses, ranging from 40 to over 300
prospectuses per topic. It also indicates the years when the prospectuses were active or when
there was a spike in activity. 
The third column identifies the key financial institutions and roles in the supply chains; 
we focus on the issuer and originator roles. 
This column also identifies all cases of financial stress and failure.

\begin{table}[hb]
\begin{small}
\centering
\begin{tabular} {|l|l|l|c|c|}
\hline
\hline
{Topic} & {DTM Topic Type} & {Supply Chain} & {Toxic} & {Model}\\
\hline
\hline
{Topic 3} & {Dynamic DTM topic.} & {Ameriquest and Weyerhauser and PHH are the issuers.} & {} & {}\\
{} & {41 prospectuses.} & {} & {Yes} & {$\uparrow$ $\uparrow$}\\
{} & {Active all years.} & {{\bf Ameriquest failure in 2007. Weyerhauser was prosecuted.}} & {} & {}\\
\hline
{Topic 7} & {Stable DTM topic.} & {IndyMac is the issuer and is also in many roles.} & {} & {}\\
{} & {92 prospectuses.} & {} & {Yes} & {$\uparrow$ $\uparrow$}\\
{} & {Active in 2006, 2007.} & {{\bf IndyMac failure in 2008.}} & {} & {}\\
\hline
{Topic 26} & {Dynamic DTM topic.} & {Fremont and American Home are the originators.} & {} & {}\\
{} & {85 prospectuses.} & {Fremont and Fieldstone and Ace Securities} & {Yes} & {$\uparrow$ $\uparrow$}\\
{} & { } & {and Deutsche Bank are the issuers.} & {} & {}\\
{} & { Peak in 2006.} & {{\bf American Home, Fieldstone failures in 2007 and 2008, resp.}} & {} & {}\\
\hline
{Topic 27} & {Stable DTM topic.} & {Structured Asset and Lehman are the issuers.} & {} & {}\\
{} & {144 prospectuses.} & {Aurora is an originator.} & {Yes} & {$\uparrow$ $\uparrow$}\\
{} & {Peak in 2005.} & {{\bf Lehman failure in 2008 and Aurora failure in 2012.}} & {} & {}\\
\hline
\hline
{Topic 8} & {Stable DTM topic.} & {Wilmington Trust is issuer. Countrywide in many roles.} & {} & {}\\
{} & {192 prospectuses.} & {Argent and Olympus are originators.} & {Partial} & {$\uparrow$ $\uparrow$}\\
{} & {Peak in 2006.} & {{\bf Argent failure in 2007.}} & {} & {}\\
\hline
{Topic 10} & {Dynamic DTM topic.} & {GMAC is the issuer. Encore Credit Corp. is originator.} & {} & {}\\
{} & {45 prospectuses.} & {GMAC faced financial difficulty and re-branding in 2009.} & {Partial} & {$\uparrow$ -}\\
{} & {Peak in 2003.} & {{\bf Encore was purchased by Bear Stearns, which failed in 2008.}} & {} & {}\\
\hline
{Topic 18} & {Stable DTM topic.} & {Morgan Stanley is the issuer and in many roles.} & {} & {}\\
{} & {80 prospectuses.} & {Long Beach is originator; owned by Washington Mutual.} & {Partial} & {$\uparrow$ $\uparrow$}\\
{} & {} & {Flagstar Bank is originator; very active in the subprime market.} & {} & {}\\
{} & {Peak in 2006.} & {{\bf Washington Mutual failure in 2009.}} & {} & {}\\
\hline
\hline
\end{tabular}
\end{small}
\caption{Summary of topics that are labeled toxic or partially toxic and increase risk of failure in the comprehensive model.}
\label{tab:mixtopics1}
\end{table}

\begin{table}[hb]
\begin{small}
\centering
\begin{tabular} {|l|l|l|c|c|}
\hline
\hline
{Topic} & {DTM Topic Type} & {Supply Chain} & {Toxic} & {Model}\\
\hline
\hline
{Topic 9} & {Dynamic DTM topic.} & {Merrill Lynch in multiple roles.} & {} & {}\\
{} & {107 prospectuses.} & {American Home Mortgage is an issuer.} & {Partial} & {$-$ $-$}\\
{} & {Active across all years.} & {{\bf American Home failure in 2007.}} & {} & {}\\
\hline
{Topic 12} & {Evolving DTM topic.} & {Bear Stearns is the issuer.} & {} & {}\\
{} & {167 prospectuses.} & {EMC is an originator; other roles.} & {Yes} & {$-$ $-$}\\
{} & {} & {{\bf Bear Stearns failure in 2008.}} & {} & {}\\
\hline
{Topic 14} & {Stable DTM topic.} & {Large topic. Countrywide in many roles.} & {} & {}\\
{} & {307 prospectuses} & {Homecomings Financial (GMAC) is an originator.} & {Partial} & {$-$ $-$}\\
{} & {Active all years.} & {{\bf Homecomings Financial failure in 2008.}} & {} & {} \\
\hline 
{Topic 25} & {Dynamic DTM topic.} & {Maia Mortgage Finance is the issuer.} & {} & {}\\
{} & {131 prospectuses.} & {} & {Partial} & {$-$ $-$}\\
{} & {Peak in 2006.} & {{\bf Financial difficulties in 2009.}} & {} & {}\\
\hline
\hline
{Topic 11} & {Stable DTM topic.} & {Large topic. Banc of America in many roles.} & {} & {}\\
{} & {125 prospectuses.} & {} & {Partial} & {$-$ $\downarrow$}\\
{} & {Active all years.} & {{\bf Banc of America received TARP funding.}} & {} & {} \\
\hline
{Topic 20} & {Stable DTM topic.} & {Small topic. Wells Fargo dominant in many roles.} & {} & {}\\
{} & {40 prospectuses.} & {} & {Partial} & {$\downarrow$ $\uparrow$}\\
{} & {} & {{\bf EMC Mortgage failure. Equity One failure.}} & {} & {} \\
\hline
{Topic 22} & {Stable DTM topic.} & {Small topic. First Horizon in many roles.} & {} & {}\\
{} & {40 prospectuses.} & {} & {Partial} & {- $\downarrow$}\\
{} & {} & {{\bf First Horizon received TARP funding in 2008.}} & {} & {} \\
\hline
\hline
\end{tabular}
\end{small}
\caption{Summary of topics that are labeled partially toxic and have a nuanced impact in the comprehensive model.}
\label{tab:mixtopics2}
\end{table}

The findings for these fourteen toxic or partially toxic topics from the comprehensive model
is as follows: 
Four toxic topics (3, 7, 26 and 27 in Table \ref{tab:mixtopics1}) 
increase the risk of failure for both FE and FNE in the comprehensive model. 
Three partially toxic topics (8, 10 and 18 in Table \ref{tab:mixtopics1}) 
similarly increase the risk of failure for 
both FE and FNE in the comprehensive model.

One toxic topic (12) and three partially toxic topics (9, 14 and 25), 
summarized in Table \ref{tab:mixtopics2}, are not significant in the comprehensive model. 
Further, three partially toxic topics (11, 20 and 22), also summarized in Table \ref{tab:mixtopics2}, 
have mixed significance; 
they are not significant in one model but slightly lower the risk of failure in the other. 
This group of seven topics provides more nuanced findings as discussed next.

Consider Topic 14, with over 300 prospectuses and almost 4,000 securities; 
it is the largest community that we identified. 
\texttt{ Countrywide}, a major player in the \texttt{resMBS} market, dominates this community and plays multiple roles. 
\texttt{Homecomings Financial (GMAC)}, a toxic financial institution, originated some mortgages. 
This topic was active and issued prospectuses across all years. 
We posit that in this large and partially toxic topic, not all of the prospectuses and securities may have been
equally impacted by the toxic supply chain. 
{\em Potentially, there is some subset of prospectuses and securities that are directly 
associated with the failed \texttt{Homecomings Financial (GMAC)}, and that were issued in the 
throes of the housing bubble, where this subset may experience a higher risk of failure.}
However, since this subset of securities are merged in this topic within a very large group of securities, 
the toxic performance of the subset may be somewhat diluted. 
Thus, despite Topic 14 being partially toxic, it was not significant in the comprehensive model.

Finally, we consider the toxic Topic 12 and the two partially toxic Topic 20 and 22. 
They are all small(er) topics, each with approximately 600+ securities. 
They are closely associated with multiple toxic financial institutions
(\texttt{Bear Stearns}, \texttt{GMAC}, \texttt{EMC Mortgage}, \texttt{Equity One}). 
We posit that the security and prospectus level features, 
including the initial ratings for these securities, 
may have well captured the increased risk of failure. 
Consider the following observations:
\begin{itemize}
\item
Topic 20 and 22 had the highest fraction of \_B\_ securities issued by any topic, 44\% and 38\%, respectively.
The sample average was 24\%. 
\item
Over 34\% of the securities issued by Topic 12 are \_B\_ securities; this too is well above the sample average
of 24\%. 
\item
Topic 20 had the highest percentage of \_A\_  securities with the SSUP label, at 33\%;
the sample average was 13\%.
\end{itemize}

The security and prospectus level features above
very likely captured the increased risk associated with these topics.
Combined with the small size of the topics, 
the topics itself may not have added significant additional risk of failure.
Consequently, the topics were not significant in the Level 3 model. 
Nonetheless, we posit that these three toxic topics, and their securities, 
remain prime candidates for poor performance and higher risk of failure.

To summarize, we observe that many toxic and partially toxic topics are associated with an 
increased risk of failure, even after accounting for security and prospectus level characteristics.
In other words, the securities produced by these communities are riskier
than we would expect when looking at the Level 1 and Level 2 features. 
On the other hand, for
some toxic and partially toxic topics that are not significant in the comprehensive model, 
the Level 1 and Level 2 features of the securities capture the increased risk of failure.


\section{Summary, Generalizability and Future Research}

This study is based on the novel \texttt{resMBS} dataset, which extracted financial 
institutions and the roles that they played in the supply chain from prospectuses.
It is the first to examine how financial institutions formed communities along
the supply chain and evaluate the impact of the communities on the financial performance of securities.
We present three complementary models to gain insights into the drivers of performance.
The models examine first security level and then prospectus level features.
Finally, a comprehensive model introduces features to represent communities. 
The models reveal several insights. First, several security level features are drivers of financial performance. 
Next, prospectus level features such as the presence of \texttt{SSUP} securities can have an impact. 
Finally, the supply chains represented by the topics have additional impact.

This study thus illustrates that identifying topics with poor financial performance 
and toxic topics is a good first step. 
However, we argue that the three levels of models are needed to fully understand the 
impact of toxic communities and the supply chain.

Using a variety of real-world evidence, we identify institutions that were active in 
the subprime market and failed or faced penalties during or after the crisis. 
We then compare the financial performance of the topics vis-à-vis the significance of 
those topics in our comprehensive model. 
We demonstrate that toxic topics increase the risk of failure, while non-toxic topics 
decrease the risk of failure. 
This finding is noteworthy, as no financial information was used when building the topics. 

We briefly discuss the generalizability of our research. 
Our novel data science contributions lie in {\em using relationships such as a financial 
supply chain to construct models of financial outcomes}. 
There are many potential dataset contexts where this approach can be applied, 
such as the securitization of student loans or the issuance of prospectuses for corporate bonds. 
In the corporate bond prospectus context, contingencies (conditional outcomes that depend on 
triggering events) are very relevant to the model of financial performance. 
The features representing the financial supply chain (financial institutions and roles for the 
\texttt{resMBS dataset}) would have to be augmented with features representing the contingencies 
in corporate bond prospectuses. 
More recently, there has been significant interest in understanding the exposure of financial 
contracts to climate change-related triggering events. 
In this scenario, additional features capturing potentially complex ownership relationships 
will augment the information on the financial supply chain.
Similar to the temporal evolution of the financial supply chain, 
mergers and acquisitions may change the ownership relationships. 
We note that in all of these cases, while the data may be available to the public, 
creating and curating the dataset would require significant effort.

Our methodological contribution is the synergistic application of machine learning 
with text corpora, e.g., dynamic topic modeling, together with more traditional 
analytical modeling, e.g., logistic regression. 
A very successful similar approach has been applied to the Business Open Knowledge Network (BOKN) 
dataset \cite{Pujara2021}. 
In this project, language embeddings are constructed over the Web text of company websites 
and temporal evolution is captured using data from archival sources such as the 
Internet Archive / Wayback Machine. 
The project builds on our use of topic models and generates more sophisticated 
embeddings such as Doc2Vec and RoBERTa \citep{devlin2019}. 
Similar to the way we model the financial supply, the embeddings of the BOKN data are used 
to construct competitor networks for companies. 
These corporate competitor networks are then used for multiple downstream tasks 
including (i) predicting the profitability of (public) companies; 
(ii) predicting the industry sector of companies; 
and (iii) predicting the reported actual competitors of companies. 
Both unsupervised and supervised approaches (Siamese networks) are used for these 
prediction tasks. Preliminary results indicate that the language embeddings over 
the Web text can indeed play a successful role in constructing competitor networks and 
making accurate predictions.

Finally, we note additional datasets where there is overlap with our data science 
and / or methodological approaches. 
This includes an IBM project to create financial networks 
\citep{burdick2011deb,hernandez2010unleashing} 
and the FEIII 2017 Challenge Dataset, which extracted relationships between financial firms 
from their filings with the SEC \citep{Raschid2017}.

While the volume of private label residential mortgage backed securities issued in the 
US decreased significantly after 2007, these products have seen a recent resurgence, 
which is a motivation for future research.
In this light, multiple interesting research trajectories can build on our work. 
The first is prediction. While the models in this paper rely on postmortem labels, 
the payment history for these securities, including various shortfalls and 
delayed payments, is also gathered on a real time basis. 
The availability of such contemporaneous data could result in the development of 
prediction models or time to failure models. 
The next possible research direction is the design of a monitoring framework for financial products. 
The \texttt{resMBS} dataset presents interesting temporal dynamics across multiple dimensions.
There is constant change of financial institutions and roles across the supply chain. 
We observe an evolution in prospectus composition, and anecdotal research indicates diminishing 
quality control over the ratings of these products. Outcomes also change over time.
An ability to account for these dynamics needs to be at the heart of any monitoring system.

Our research combines multiple machine learning approaches to study the 
supply chain and its impact on financial performance of\texttt{resMBS} securities. 
While we do not claim to identify causal links, 
our analysis demonstrates the power of utilizing financial big data
and computational methods to understand complex financial products.

\ACKNOWLEDGMENT{%
We thank the following individuals: 
Nancy Wallace and Paolo Issler (University of California Haas School of Business)
and Joe Langsam for their help in identifying the \texttt{resMBS} prospectuses and 
labeling the performance of these securities; 
Doug Burdick and Rajasekar Krishnamurthy (IBM Research) for supporting the text 
extraction task using the IBM SystemT platform; 
Soham De and Minchao Shao for implementing the \texttt{resMBS} extraction pipeline;
Zheng Xu, Elena Zotkina, Aaron Hunt, Chi-Hung Chen and Prabhath Kollimarla
for technical support with data cleaning, data integration and topic model 
based analytics. 
This research was partially supported by 
NSF grants CNS1305368 and NIST award 70NANB15H194.

}

\clearpage
\newpage
\begin{APPENDIX}{}

\section{Brief Introduction to Topic Modeling}

LDA is a statistical model that aims to explain a set of documents using unobserved topics.
LDA is based on a generative statistical model for collections of discrete data (documents);
it allows us to extract topics based on assumptions about the probability distributions
that were used to generate the documents.
At a high level, it is assumed that to create a document, one first randomly chooses a
distribution over a collection of (unobserved) topics. Then, for each word in a document,
one independently and randomly chooses a topic from the previously sampled mix of topics
assigned to the document, then again independently and randomly draws a word from the
(unobserved) word distribution for that topic.
Based on these assumptions about the generative mechanism and the associated underlying
probability distributions, topic models aim to extract the underlying topic structure via
maximum likelihood estimation.

More specifically, let $P( z )$ represent the distribution for a topic $z$ from $K$ topics
in a particular document.
Let $P( w | z )$ represent the probability distribution over words $w$ for the given topic $z$.
Let $P( z_i = j )$ be the probability that the $j$-th topic was sampled for the
$i$-th word token in the document.
Let $P( w_i | z_i = j )$ be the probability of word $w_i$ for topic $j$.
We then have the following probability distribution for words within a document:

\begin{eqnarray}
P ( w_i ) = \sum_ {j = 1} ^{K} P ( w_i | z_i = j ) \times P ( z_i = j )
\end{eqnarray}

LDA represents each document as a random mix over latent topics.
Each topic is characterized by a distribution over words.
Associated with these two distributions are the hyperparameters $\ga$ and $\gb$,
corpus level parameters assumed to be sampled once in the process of generating a corpus.
The variable $\gt$ is a document level variable (a topic mixture over $K$ topics,
sampled from a Dirichlet distribution parameterized by $\ga$) sampled once per document.
The variables $z_n$ and $w_n$ are word level variables sampled once for each word in each document.
Specifically, for each word, a topic $z_n$ is sampled from a multinomial distribution
conditioned on $\gt$, and each word $w_n$ is sampled from a multinomial distribution
conditioned on $z_n$ and parameterized by $\gb$.

One can re-write the probability of a document with $N$ words as follows:
\begin{eqnarray}
p(\w | \ga, \gb) = \int p(\gt | \ga)(\prod_{n=1}^N \sum_{z_n} p(z_n | \gt) p(w_n | z_n,\gb)) d\gt
\end{eqnarray}

where $\w=\{w_1, w_2, \ldots, w_N\}$ is a set of $N$ words. Figure \ref{fig:LDA} is a graphical representation of the LDA process for a collection of $M$ documents.

\begin{figure}[hbtp]
\centerline{
\includegraphics[width=0.6\linewidth]{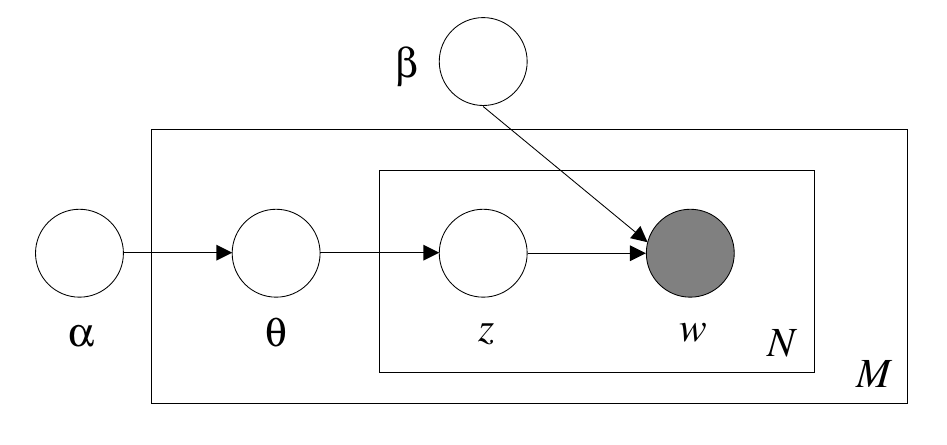}
}
\begin{center}
\caption{A graphical representation of LDA for a text corpus (~\cite{BNJ2003}).
The boxes are "plates" representing replicates,
where the outer plate represents M documents and the inner plate represents
the repeated choice of topics and words within a document. }
\label{fig:LDA}
\end{center}
\end{figure}

The LDA process just described assumes that the documents are drawn from a static set of topics.
Extensions that relax this assumption include a continuous non-Markovian model of topics over
time (ToT) (\cite{Wang2006}) and dynamic topic models (DTM) (\cite{BleiDTM2006}).
ToT assumes that each topic is associated with a continuous distribution over the time slices.
This allows a modeler to model how some topics may be more popular over certain time periods.
In that case, for each document, the mix of topics ($\theta$ in LDA) is influenced by both the
word co-occurrence and the timestamps.
In the DTM extension, which is the approach we adopt, the document
collection is also divided by time, e.g., in our case a time slice for each year.
As the Dirichlet distribution does not lend itself well to sequential modeling, the time
dependent parameters evolve over time according to a Brownian motion.
For example, $\beta_{t,k}$ can be expressed as follows:

\begin{eqnarray}
\gb_{t,k} | \gb_{t-1,k} \sim \mathcal{N}(\gb_{t-1,k},\sigma^2 I).
\end{eqnarray}

The evolution of $\ga$ and $\gt$ are expressed in a similar manner.
We point the interested reader to \cite{BleiDTM2006} for additional technical details.
Figure \ref{fig:DTM} is a graphical representation of the DTM process.
Through the evolution described above, the topics can evolve over time, with the speed of evolution controlled by a tuning parameter.
Setting the parameter at one extreme will convert DTM to a static topic model,
while setting it at the other extreme will result in independent topics in each time slice.

\begin{figure}[hbtp]
\centerline{
\includegraphics[width=0.6\linewidth]{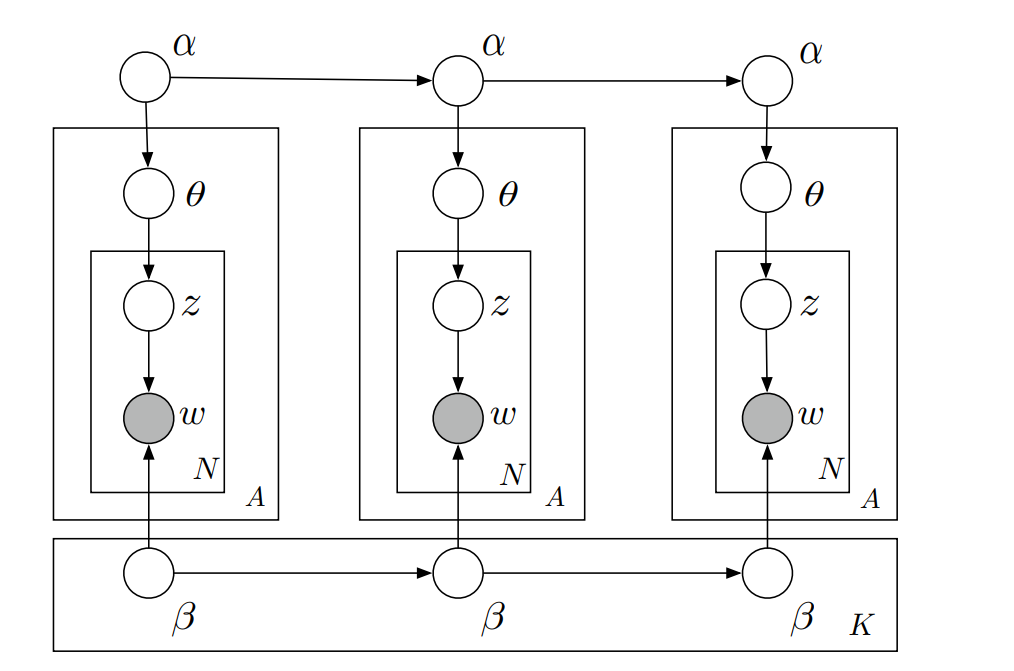}
}
\begin{center}
\caption{A graphical representation of a dynamic topic model (DTM) for three time slices (~\cite{BleiDTM2006}).
The parameters for each topic, $\gb_{t,k}$, as well as the parameter for the topic mix $\ga_{t}$, evolve over time.}
\label{fig:DTM}
\end{center}
\end{figure}

\section{Data Details and Summary Statistics} 

\begin{table}[hb]
\begin{small}
\begin{tabular} {|l|r|r|r|r|}
\hline
\hline
{Role} & {Frequency} & {\%} & {Frequency} & {\%} \\
{} & {maximum: 4787} & {} & {maximum: 3537} & {} \\ \hline
Issuer & 3676 & 76 & 3183 & 89\\
Originator & 1757 & 36 & 1713 & 48\\
Seller & 3131 & 65 & 2588 & 73\\
Trustee & 3722 & 77 & 3284 & 92\\
{Servicer (many variants)} & 4335 & 90 & 3502 & 98\\
Depositor & 4081 & 85 & 3314 & 93\\
Sponsor & 1731 & 36 & 1651 & 46\\
{Securities administrator} & 873 & 18 & 842 & 23\\
Custodian & 733 & 15 & 706 & 19\\
{Swap counterparty} & 516 & 10 & 507 & 14\\
{Cap counterparty} & 424 & 8 & 392 & 11\\
Insurer & 355 & 7 & 330 & 9\\
Underwriter & 228 & 4 & 143 & 4\\
\hline
\hline
\end{tabular}
\end{small}
\caption{Distribution of roles across prospectuses; 
4,787 prospectuses (total) and 3,537 prospectuses (with Bloomberg data).}
\label{tab:summary2}
\end{table}

\begin{table}[tb]
\begin{small}
\begin{tabular} {|p{0.3\linewidth}|p{0.7\linewidth}|}
\hline
\hline
{Attribute} & {Description}\\\hline
{CUSIP:} & {Unique identifier.}\\
{Name:} & {Unique name.}\\
{Tranche Description:} &
{Essential characteristics of multi-class mortgage and asset-backed securities (CMO, ABS, CMBS).}\\
{Original Principal:} &
{The principal balance at issuance of the security.}\\
{Maturity:} &
{Date the principal of a security is due and payable.}\\
{Count of Loans:} &
{The current number of loans, created as collateral for the deal, which are still outstanding.}\\
{60+ DQ:} &
{Percentage of loans which are 60 or more days delinquent, including loans in foreclosure, bankruptcy.}\\
{Cumulative shortfall:} &
{Cumulative supported shortfall that has yet to be repaid.}\\
{2 shortfall:} &
{Cumulative shortfall that will not be repaid.}\\
{Cumulative Loss:} &
{Cumulative writedown on the principal.}\\
{Historic Cashflow:} &
{A complete set of known historical cashflows starting from the bond issuance.
The data includes the period number, date, coupon, interest, 
principal paid, and principal balance.}\\
{Ratings:} &
{Multiple fields with the original and current ratings from Moody, Fitch, and SP;
corresponding dates and a composite rating from Bloomberg L.P.}\\
\hline
\hline
\end{tabular}
\end{small}
\caption{Attributes for securities (\cite{BloombergLP}).}
\label{tab:attributes}
\end{table}

\begin{table}[hb]
\begin{small}
\begin{tabular} {|l|r|r|r|r|r|r|r|r|}
\hline
\hline
{} & {2002} & {2003} & {2004} & {2005} & {2006} & {2007} & {2008} & {Total}\\
\hline
{Prospectus count (initial) } & 452 & 640 & 881 & 1047 & 1019 & 667 & 81 & 4787\\ \hline
{With issuer or originator} & 231 & 398 & 577 & 744 & 923 & 593 & 71 & 3537\\ 
{and match in Bloomberg} & & & & & & & & \\ \hline
{With $>=$ 3 DISTINCT FE mentions} & \multicolumn{7}{c|}{} & {3146}\\
{} & \multicolumn{7}{c|}{} & {}\\ \hline
{With $>=$ 5 (Role, FE) mentions} & \multicolumn{7}{c|}{} & {4472}\\
{} & \multicolumn{7}{c|}{} & {}\\ \hline
\hline
\end{tabular}
\end{small}
\caption{Count of Prospectuses: 
with issuer and/or originator identified (4,787); with match in Bloomberg (3,537);
further filtered for specific models.}
\label{tab:summary1}
\end{table}

\begin{table}[hb]
\begin{small}
\begin{tabular} {|l| l|}
\hline
\hline
{} & {\texttt{resMBS}}\\
\hline
{Count of documents} & {4472} \\
{Count of distinct financial institutions} & {85} \\
{Count of distinct roles} & {27} \\
{Distinct (Role\_FI) pairs} & {267} \\
{Count of (Role\_FI) occurrences} & {41075} \\
\hline
\hline
\end{tabular}
\end{small}
\caption{Summary statistics for \texttt{resMBS} topic model experiments.}
\label{tab:summary4}
\end{table}

\begin{table}[hb]
\begin{small}
\begin{tabular} {|p{0.2\linewidth}|l|p{0.1\linewidth}|p{0.5\linewidth}|}
\hline
\hline
{Features} & {Level} & {Number of Features} & {Description} \\
\hline
Moody’s initial rating & Security & 11 & Over 60 initial rating values are grouped together into 9 higher level rating levels, in addition to a value for ``not rated'' and ``no value.''\\
Payoff characteristics & Security & 73 & The payoff structure of a security is captured with 73 binary indicators.\\
Initial mortgage amount & Security & 1 & The initial mortgage amount of the pool of mortgages serving the security.\\
Security class & Security & 3& The class of the security, \_A\_, \_M\_, or \_B\_.\\
Presence of SSUP & Prospectus & 1 & A binary indicator for whether any security within a prospectus is a \texttt{SSUP} security.\\
Class distribution in a prospectus & Prospectus & 6 & Six variables that capture the fraction of the number securities within each class ( \_A\_, \_M\_, or \_B\_) and the volume (in USD) within the same.\\
Financial communities & Prospectus & 30 & Binary indicator variables for the topic that each prospectus is assigned to using the largest topic weight.\\
Annual controls & Security & 6 & Binary indicators for the year of issuance.\\
\hline
\hline
\end{tabular}
\end{small}
\caption{Overview of the features used in the study }
\label{tab:datasummary}
\end{table}

\clearpage

\section{Detailed LASSO Model Results}

\begin{table}[hb]
\begin{small}
\centering
\begin{tabular} {|l| l| l| l| l| }
\multicolumn{5}{c}{All Securities}\\
\hline\hline
Model & \multicolumn{2}{c|}{Accuracy}& \multicolumn{2}{c|}{F1-Score}\\
\hline
& FE & FNE & FE & FNE \\
Security Level Model & 91.3\% & 84.7\% & 0.891 & 0.857 \\
Prospectus Level Model & 91.2\% & 85.0\% & 0.889 & 0.860 \\
Comprehensive Community Level Model & 91.6\% & 85.4\% & 0.894 & 0.864 \\
\hline \hline
\multicolumn{5}{c}{Class \_A\_}\\
\hline\hline
Model & \multicolumn{2}{c|}{Accuracy}& \multicolumn{2}{c|}{F1-Score}\\
\hline
& FE & FNE & FE & FNE \\
Security Level Model & 95.8\% & 82.6\% & 0.734 & 0.623 \\
Prospectus Level Model & 95.7\% & 82.7\% & 0.732 & 0.639 \\
Comprehensive Community Level Model & 95.9\% & 82.6\% & 0.737 & 0.644 \\
\hline \hline
\multicolumn{5}{c}{Class \_M\_}\\
\hline\hline
Model & \multicolumn{2}{c|}{Accuracy}& \multicolumn{2}{c|}{F1-Score}\\
\hline
& FE & FNE & FE & FNE \\
Security Level Model & 86.0\% & 85.0\% & 0.903 & 0.909\\
Prospectus Level Model & 86.2\% & 86.0\% & 0.904 & 0.914\\
Comprehensive Community Level Model & 86.5\% & 86.2\% & 0.906 & 0.916\\
\hline \hline
\multicolumn{5}{c}{Class \_B\_}\\
\hline\hline
Model & \multicolumn{2}{c|}{Accuracy}& \multicolumn{2}{c|}{F1-Score}\\
\hline
& FE & FNE & FE & FNE \\
Security Level Model & 87.2\% & 88.0\% & 0.912 & 0.930 \\
Prospectus Level Model & 86.5\% & 88.1\% & 0.907 & 0.930 \\
Comprehensive Community Level Model & 87.0\% & 89.3\% & 0.910 & 0.937 \\
\hline \hline
\end{tabular}
\end{small}
\caption{Performance metrics for the three LASSO models, for the entire dataset, and for each security class,
\_A\_, \_M\_ and \_B\_.}
\label{tab:performance}
\end{table}

\begin{table}[hb]
\begin{small}
\centering
\begin{tabular} {|l| l| l|c| l| l| l| }
\hhline{===~===}
Variable & FE & FNE && Variable & FE & FNE\\
\cline{1-3}
\cline{5-7}
Intercept & -0.717 & 0.933 && \multicolumn{3}{c|}{Moody's Initial Rating}\\
\cline{1-3} \cline{5-7}
\multicolumn{3}{|c|}{Class \& Amount} && A & 0.198 &  \\
\cline{1-3}
MTG.ORIG.AMT & -0.003 & && Aa & -0.599 & -0.53 \\
IsA & -4.669 & -3.307 && Aaa & -0.252 & -0.063 \\
IsB & 0.079 & 0.046 && B & 0.144 & 0.005 \\
\cline{1-3}
\multicolumn{3}{|c|}{Tranche Type} && Ba & 0.81 & 0.735 \\
\cline{1-3}
AD & -0.449 & -0.014 && Baa & 0.517 & 0.572 \\
AFC & -0.248 & 0.305 && C & 1.827 & 0.065 \\
AS & 0.223 & 0.705 && Not rated & 0.516 & 0.319 \\
\cline{5-7}
CMPLX & -0.198 & && \multicolumn{3}{c|}{Annual Controls}\\
\cline{5-7}
CPT & 0.98 & 0.88 && 2002 & -4.518 & -3.297 \\
CSTR & 0.31 & 0.65 && 2003 & -3.15 & -2.119 \\
DLY & 0.602 & 0.066 && 2004 & -1.673 & -1.292 \\
EXCH & -0.194 & 0.666 && 2006 & 2.08 & 1.291 \\
EXE & -1.468 & && 2007 & 2.113 & 1.364 \\
\hhline{|~|~|~|~===}
FLT & -0.093 & -0.135\\ 
FTV & -0.336 & \\
INV & -0.145 & \\
IRC & -0.06 & -0.064 \\
MEZ & 0.971 & \\
MR & -0.543 & \\
NAS & 1.019 & 0.698 \\
NTL & -0.311 & -0.519 \\
OC & -3.215 & \\
PAC1 & -0.512 & -0.24 \\
PT & 0.056 & \\
R & 0.139 & \\
RAKE & -0.508 & -0.303 \\
RSTP & -0.071 & \\
RTL & & 1.83 \\
SC & -3.208 & \\
SEQ & -1.243 & -0.35 \\
SSNR & -1.2 & 0.294 \\
SSUP & 4.413 & 2.188 \\
STEP & -0.106 & \\
SUB & 1.433 & 0.496 \\
SUP & & -0.074 \\
TAC.22. & -0.391 &\\
W & -0.289 & \\
Z & & 1.486 \\
\hhline{===~~~~}

\end{tabular}
\end{small}
\caption{The regression coefficients for the Security Level Model.
Blank cells indicate that the variable was not retained by the FE or FNE model. 
Note that variables that are not retained by either the FE or FNE model are not included.}
\label{tab:model1coeff}
\end{table}

\begin{table}[hb]
\begin{small}
\centering
\begin{tabular} {|l| l| l|c| l| l| l| }
\hhline{===~===}
Variable & FE & FNE && Variable & FE & FNE\\
\cline{1-3}
\cline{5-7}
Intercept & -0.635 & 1.260 && \multicolumn{3}{c|}{Moody's Initial Rating} \\
\cline{1-3} \cline{5-7} 
\multicolumn{3}{|c|}{Class \& Amount} && A &	0.207	&	0.025\\
\cline{1-3}
MTG.ORIG.AMT&	-0.00309	&		-0.00022	&& Aa &	-0.611	&	-0.595	\\
IsA	&	-4.658	&	-3.439	&&	Aaa	&	-0.320	&	-0.050	\\
IsB	&	0.045	&	0.028	&&	B	&	0.168	&	0.218	\\
\cline{1-3}											
\multicolumn{3}{|c|}{Tranche Type} && 						Ba	&	0.863	&	0.932	\\
\cline{1-3}											
AD	&	-0.438	&	-0.136	&&	Baa	&	0.554	&	0.682	\\
AFC	&	-0.309	&	0.343	&&	C	&	1.932	&	0.513	\\
AS	&	0.062	&	0.665	&&	Not rated	&	0.457	&	0.294	\\
\cline{5-7}											
CMPLX	&	-0.149	&	-0.023	&&	\multicolumn{3}{c|}{Prospectus Characteristics}\\					
\cline{5-7}											
CPT	&	0.946	&	1.216	&&	Frac A	&	-0.729	&	-0.748	\\
CSTR	&	0.249	&	0.608	&&	HasSSUP	&	0.573	&	0.538	\\
DGT	&		&	-0.279	&&	Vol A	&	0.282	&	0.070	\\
DLY	&	0.622	&	0.157	&&	Vol M	&	-0.487	&	-0.449	\\
\cline{5-7}											
EXCH	&	-0.107	&	0.759	 && \multicolumn{3}{c|}{Annual Controls}\\						
\cline{5-7}											
EXE	&	-1.262	&		&&	2002	&	-4.367	&	-3.365	\\
FLT	&	-0.238	&	-0.270	&&	2003	&	-3.081	&	-2.201	\\
FTV	&	-0.500	&	-0.152	&&	2004	&	-1.647	&	-1.379	\\
INV	&	-0.171	&		&&	2005	&	2.047	&	1.340	\\
IRC	&	-0.063	&	-0.091	&&	2006	&	2.017	&	1.381	\\
\hhline{|~|~|~|~===}											
MEZ	&	0.987	&		\\						
MR	&	-0.622	&		\\						
NAS	&	0.824	&	0.587	\\						
NTL	&	-0.579	&	-0.838	\\						
OC	&	-3.307	&		\\						
PAC1	&	-0.486	&	-0.436	\\						
PIP	&		&	-0.091	\\						
PT	&	0.050	&	-0.141	\\						
RAKE	&	-0.636	&	-0.975	\\						
RSTP	&	-0.309	&	-0.229	\\						
RTL	&		&	2.225	\\						
SC	&	-3.198	&		\\						
SEQ	&	-1.272	&	-0.527	\\						
SSNR	&	-1.315	&	0.115	\\						
SSUP	&	4.196	&	2.037	\\						
STEP	&	-0.016	&		\\						
SUB	&	1.386	&	0.477	\\						
SUP	&		&	-0.307	\\						
TAC.1.22.	&		&	0.338	\\						
TAC.11.	&		&	-0.041	\\						
TAC.2.22.	&		&	-0.096	\\						
TAC.22.	&	-0.434	&		\\						
W	&	-0.231	&		\\						
Z	&		&	1.636	\\						
\hhline{===~~~~}
\end{tabular}
\end{small}
\caption{
The regression coefficients for the Prospectus Level Model.
Blank cells indicate that the variable was not retained by the FE or FNE model.
Note that variables that are not retained by either the FE or FNE model are not included.}
\label{tab:model2coeff}
\end{table}

\begin{table}[hb]
\begin{small}
\centering
\begin{tabular} {|l| l| l|c| l| l| l| }
\hhline{===~===}
Variable & FE & FNE && Variable & FE & FNE\\
\cline{1-3}
\cline{5-7}
Intercept & -0.454 & 1.368 && \multicolumn{3}{c|}{Moody's Initial Rating}\\ 
\cline{1-3} \cline{5-7} 
\multicolumn{3}{|c|}{Class \& Amount} && A & 0.188 & 0.01 \\
\cline{1-3}
MTG.ORIG.AMT & -0.003 &-6.02E-05 && Aa & -0.627 & -0.613 \\
IsA & -4.9 & -3.513 && Aaa & -0.43 & -0.072 \\
IsB & 0.233 & 0.179 && B & 0.157 & 0.175 \\
\cline{1-3}
\multicolumn{3}{|c|}{Tranche Type} && Ba & 0.845 & 0.917 \\
\cline{1-3}
AD & -0.32 & -0.031 && Baa & 0.538 & 0.664 \\
AFC & -0.17 & 0.388 && C & 1.861 & 0.664 \\
AS & & 0.588 && Not rated & 0.462 & 0.266 \\
\cline{5-7}
CMPLX & -0.013 & && \multicolumn{3}{c|}{Prospectus Characteristics}\\ 
\cline{5-7} 
CPT & 0.858 & 1.221 && FracA & -0.471 & -0.56 \\
CSTR & 0.407 & 0.698 && HasSSUP & 0.467 & 0.449 \\
DGT & & -0.004 && VolA & 0.12 & \\
DLY & 0.547 & 0.239 && VolM & -0.751 & -0.661 \\
\cline{5-7}
EXCH & -0.085 & 0.768 && \multicolumn{3}{c|}{Topics}\\ 
\cline{5-7} 
EXE & -0.803 & && Topic 1 & & -0.667 \\
FLT & -0.258 & -0.399 && Topic 10 & 0.371 & \\
FTV & -0.325 & -0.198 && Topic 11 & -0.108 & -0.247 \\
INV & -0.166 & -0.042 && Topic 13 & & 0.261 \\
IRC & -0.009 & -0.055 && Topic 15 & -1.34 & -1.585 \\
MEZ & 0.833 & && Topic 16 & -0.316 & -0.349 \\
NAS & 0.761 & 0.539 && Topic 17 & -0.285 & -0.099 \\
NTL & -0.563 & -0.961 && Topic 18 & 0.489 & 0.194 \\
OC & -3.326 & && Topic 19 & -1.076 & \\
PAC1 & -0.211 & -0.315 && Topic 2 & -0.662 & -0.061 \\
PIP & & -0.468 && Topic 20 & -0.335 & 0.027 \\
PT & 0.146 & && Topic.21 & -0.441 & \\
RAKE & -0.293 & -0.906 && Topic 22 & & -0.12 \\
RSTP & & -0.125 && Topic 24 & 0.353 & 0.082 \\
RTL & & 2.243 && Topic 25 & 0.187 & 0.055 \\
SC & -3.192 & && Topic 26 & 0.412 & 0.449 \\
SEQ & -0.929 & -0.408 && Topic 27 & 0.328 & 0.371 \\
SSNR & -1.289 & 0.093 && Topic 28 & -0.122 & -0.246 \\
SSUP & 4.317 & 2.093 && Topic 29 & -1.308 & -0.457 \\
SUB & 1.096 & 0.373 && Topic 3 & 0.636 & 0.292 \\
SUP & & -0.258 && Topic 30 & -0.249 & -0.358 \\
TAC.1.22. & & 0.032 && Topic 4 & -1.09 & -0.949 \\
TAC.11. & & -0.206 && Topic 5 & -0.242 & -0.071 \\
TAC.22. & -0.186 & && Topic 6 & & 0.032 \\
TAC.33. & & -0.018 && Topic 7 & 0.323 & 0.567 \\
Z & & 1.694 && Topic 8 & 0.974 & 0.364 \\
\cline{1-3}
\multicolumn{3}{|c|}{Annual Controls} && Topic.9 & 0.027 & \\
\cline{1-3}
\hhline{|~|~|~|~===}
2002 & -4.206 & -3.192 \\
2003 & -3.064 & -2.164 \\
2004 & -1.611 & -1.327 \\
2006 & 1.995 & 1.351 \\
2007 & 2.094 & 1.484 \\
\hhline{===~~~~}
\end{tabular}
\end{small}
\caption{The regression coefficients for the Comprehensive Community Level Model. 
Blank cells indicate that the variable was not retained by the FE or FNE model.
Note that variables that are not retained by either the FE or FNE model are not included.}
\label{tab:model3coeff}
\end{table}

\end{APPENDIX}



\clearpage
\bibliographystyle{informs2014} 
\bibliography{TM_IJoC,NER+ER}


\end{document}